\documentclass[twoside,leqno,twocolumn]{article}

\usepackage[letterpaper]{geometry}

\usepackage{siamproceedings}
\usepackage[T1]{fontenc}
\usepackage{amsfonts}
\usepackage{graphicx}
\usepackage{epstopdf}
\usepackage{enumitem}
\usepackage{algorithmic}
\ifpdf
  \DeclareGraphicsExtensions{.eps,.pdf,.png,.jpg}
\else
  \DeclareGraphicsExtensions{.eps}
\fi

\newsiamremark{remark}{Remark}
\newsiamremark{hypothesis}{Hypothesis}
\crefname{hypothesis}{Hypothesis}{Hypotheses}
\newsiamthm{claim}{Claim}

\usepackage{amsopn}
\usepackage{times}
\usepackage{helvet}
\usepackage{courier}
\usepackage{natbib}
\usepackage{caption}
\usepackage{amsmath}
\usepackage{multirow}
\usepackage{booktabs}
\usepackage{placeins}
\usepackage{float}
\usepackage{stfloats}
\usepackage{subcaption}
\usepackage{hyperref}

\usepackage{xcolor}

\newcommand{\modelFullName}{\textbf{M}ulti-view \textbf{B}ehavior-aware \textbf{Diff}usion Model for Probabilistic Utility Data Imputation}

\newcommand{\modelAbvName}{MBDiff}
\newcommand{\mOneFullName}{User Behavior Extraction}
\newcommand{\mOneAbvName}{MUBE}

\newcommand{\mTwoFullName}{Behavior-aware Conditional Diffusion Model}
\newcommand{\mTwoAbvName}{BCDiff}

\title{MBDiff: Multi-view Behavior-aware Diffusion Model \\for Probabilistic Utility Data Imputation}

\author{%
  Rongchao Xu\thanks{Department of Computer Science, Florida State University,
  Tallahassee, FL, USA (\email{rx21a@fsu.edu}, \email{lj23d@fsu.edu},
  \email{dahai.yu@fsu.edu}, \email{xl24g@fsu.edu}, \email{guang@cs.fsu.edu}).}%
  \and Lin Jiang\footnotemark[1]%
  \and Dahai Yu\footnotemark[1]%
  \and Ximiao Li\footnotemark[1]%
  \and Guang Wang\footnotemark[1]\thanks{Corresponding author.}%
}

\date{}
\begin{document}
\maketitle

\fancyfoot[R]{\scriptsize{Copyright \textcopyright\ 2026 by SIAM\\
Unauthorized reproduction of this article is prohibited}}

\begin{abstract}

Utility data (e.g., electricity, water, and gas consumption), collected by ubiquitous sensors and embedded devices, often contains substantial missing values due to various factors such as device failures and data transmission issues.
The data missingness can severely impact utility billing accuracy, hinder demand forecasting, and disrupt efficient utility supply management. As a result, utility data imputation has attracted much interest from both industry and academia. 
While many studies have attempted to address this issue, most of them rely on aggregated datasets for training, overlooking rich user behavior information, which could provide valuable insights for more accurate imputation. 
However, learning comprehensive user behavior from long-term, diverse, and incomplete utility data remains a significant challenge. Moreover, leveraging user behavior information to guide imputation is nontrivial due to the indirect nature of the correlations.
To address these challenges, we propose \modelAbvName, a \modelFullName.
\modelAbvName~incorporates two key technical components:
(i) a multi-view \mOneFullName~module that learns comprehensive user behavior from multiple perspectives, including global, local, and instance-level views; and
(ii) a behavior-aware conditional diffusion model consisting of a reference selection module and a conditional attentional denoising network to impute utility data in a computationally efficient manner.
We implement and evaluate \modelAbvName~by collaborating with one of the largest municipal utility providers in Florida. Experimental results demonstrate our proposed \modelAbvName~effectively outperforms state-of-the-art baselines, e.g., it improves 7.04\% and 29.1\% on the electricity and water usage datasets for block missingness imputation, respectively.
\end{abstract}

\section{Introduction}

Missing values are common in utility data (e.g., electricity load, water usage, and gas consumption) due to various factors such as sensor malfunctions, extreme weather events (e.g., hurricanes)~\cite{li2025typhoformer, shen2025learning}, and irregular sampling, which can significantly affect real-world utility services in different aspects including billing, resource allocation, and demand-supply rebalancing. According to our collaboration with one of the largest utility providers in Florida, missing utility records during residential transitions present considerable challenges for both providers and customers. For instance, inaccurate or incomplete data may lead to unfair billing, where outgoing residents are either undercharged or overcharged, and incoming residents may be billed for consumption they did not cause. These issues can erode customer trust, prompt disputes, and potentially lead to regulatory non-compliance.

Due to its importance, utility data imputation has attracted much interest from both academia and industry, and various approaches have been proposed to address this task. Early statistical and machine learning methods, such as interpolation, k-nearest neighbors (KNN) \citep{altman1992introduction}, and matrix factorization \citep{fang2020time}, were widely used but often struggled with long and complex sequences exhibiting intrinsic dependencies.
Recently, deep learning methods have advanced utility data imputation (e.g., RNNs \cite{yoon2018estimating, cao2018brits}, CNNs \cite{wu2023timesnet}, VAEs \citep{fortuin2020gp}, GANs \citep{brophy2023generative}, diffusion models \citep{tashiro2021csdi, liu2023pristi}), which aim to learn underlying distributions for accurate imputation. However, most existing works typically rely on aggregated data for model training, often overlooking a critical factor, i.e., user-specific behaviors. These behaviors, characterized by unique trends and habitual patterns, can offer valuable information to improve imputation accuracy.
More detailed motivation and data analysis are presented in Appendix~\ref{App:motivation}, which further illustrate diverse user behaviors across different utility types and time periods.


In this paper, we focus on user-specific utility data imputation by considering unique user behaviors. 
However, there are two major challenges to achieving this objective: (1) \textit{How to effectively learn comprehensive user behavior from long-term, dynamic, and diverse historical utility data?} (2) \textit{How to accurately impute the data with extracted user behavior in a computationally efficient way?}
First, behavioral patterns can vary significantly not only across users but also for the same user over time, making it difficult to extract comprehensive and meaningful patterns. 
Furthermore, adopting machine learning for data imputation may cause the user-specific data to be temporally discontinuous due to training and testing data splitting. For instance, data may only be available for certain days (e.g., days 1, 4, 5, 7, and 14) in the training data, resulting in large temporal gaps that further complicate behavior modeling.
A straightforward way is to represent user behavior by generating a global user embedding or a chronologically sorted latent representation. However, such representations often fail to provide adequate and contextually relevant references for imputing target records on different dates.
Secondly, leveraging the learned user behavior to efficiently guide the imputation of target data records remains a challenge due to implicit correlations and high dimensionality.
In particular, it is difficult to establish dependencies between user behavior representations and the attributes of target records (e.g., timestamps or values), as such relationships are often complex, indirect, and not explicitly observable.


To address these challenges, we propose \modelAbvName, a \modelFullName, which consists of two key components: a Multi-view User Behavior Extraction module (\mOneAbvName) and a Behavior-aware Conditional Diffusion model (\mTwoAbvName). \mOneAbvName~ comprehensively extracts users' behavior patterns from different
aspects (e.g., global, local, and instance view) based on long-term, diverse, and discontinuous historical time series data.
In \mTwoAbvName, a reference selection module is firstly introduced to identify the most informative behavior patterns, thereby reducing computational cost. Subsequently, a novel behavior-aware conditional attentional denoising network is designed to accurately impute missing values by leveraging the selected reference behaviors.

In summary, the key contributions are as follows:
\begin{itemize}
\item Motivated by real-world significance, in this paper, we address the utility data imputation problem by considering complicated user behaviors, which can benefit both utility providers and customers through accurate utility billing and resource allocation for societal impact.
\item We propose a behavior-aware utility data imputation framework called \modelAbvName, which includes a multi-view 
user behavior extraction module that learns comprehensive user behaviors from various aspects (e.g., global, local, and instance view), and a computationally efficient behavior-aware conditional diffusion model with an attentional denoising network that leverages the informative behavior to impute missing values.
\item We implement and comprehensively evaluate our proposed \modelAbvName{} on three types of real-world utility data in collaboration with a large municipal utility provider. Extensive results show that \modelAbvName{} significantly outperforms state-of-the-art baselines, e.g., it improves 7.04\% and 29.1\% on the electricity and water usage datasets for block missingness imputation, demonstrating the effectiveness of our design. More importantly, our \modelAbvName\ has helped our collaborator impute over 20 million missing values in their utility data to improve their utility services, indicating its social impact.

\end{itemize}
\section{Preliminary}
\subsection{Problem Definition}\label{sec:problem}

\noindent\textbf{Definition 1: (Utility Data)} For a utility dataset collected from a group of users $U = \{u_1, u_2, \cdots, u_N\}$, where $N$ denotes the total number of users, we define
 $D_u = \{(X_1^u, d_1^u, m_1^u), (X_2^u, d_2^u, m_2^u), \cdots, (X_{N_u}^u, d_{N_u}^u, m_{N_u}^u)\}$  as the historical utility data of user $u$. Here, $X_j^u$ represents the $j$-th data record (time series) of user $u$, and $N_i$ is the total number of data records of user $u$. The timestamp $d_j^u = (dd_j^u, md_j^u, yd_j^u)$  consists of the day of a week (\(dd_j^u \in [1,7]\)), month (\(md_j^u \in [1,12]\)), and year $yd_j^u$ of $X_j^u$. The variable $m_j^u \in \{0, 1\}$ is a binary observation mask to indicate whether a value (data point) is observed or missing in a data record, where \(m_l = 1\) represents the data point is observed and \(m_l = 0\) represents the data point is missing.

\noindent\textbf{Definition 2: (Utility Data Imputation)}
Given a historical dataset $D_{u_i}$ of user $u_i$, deterministic/probabilistic utility data imputation aims to estimate the values/distributions of the missing values in each utility data record $\tilde{X}$ at the date $\tilde{d} = (\tilde{dd}, \tilde{md}, \tilde{yd})$, where \(\tilde{dd}\), \(\tilde{md}\), and \(\tilde{yd}\) represent the day, month, and year, respectively.

\subsection{Denoising Diffusion Probabilistic Models}
\label{ss:dm}
Denoising Diffusion Probabilistic Models (DDPMs) \citep{ho2020denoising} are a class of generative models designed to learn complex data distributions through a sequence of noise-adding and denoising steps. 
It consists of a forward diffusion process and a reverse generative process.
The forward process gradually corrupts the data \(x_0 \in \mathbb{R}^d\) by adding Gaussian noise over \(T\) discrete time steps:
\begin{equation}
    q(x_t | x_{t-1}) = \mathcal{N}(x_t; \sqrt{\alpha_t} x_{t-1}, (1 - \alpha_t)I),
\end{equation}
\vspace{-10pt}
\begin{equation}
    q(x_{1:T} | x_0) = \prod_{t=1}^T q(x_t | x_{t-1}),
\end{equation}
where \(\alpha_t \in (0, 1)\) controls the variance schedule. After \(T\) steps, the data becomes nearly isotropic Gaussian noise.
The reverse process aims to iteratively reconstruct \(x_0\) from \(x_T\). This is achieved by learning a parameterized model \(p_\theta(x_{t-1} | x_t)\), which is also Gaussian:
\begin{equation}
    p_\theta(x_{t-1} | x_t) = \mathcal{N}(x_{t-1}; \mu_\theta(x_t, t), \Sigma_\theta(t)).
\end{equation}
The mean \(\mu_\theta(x_t, t)\) is predicted by a neural network trained to approximate the noise added during the forward process.
The model is trained by minimizing a simplified variational lower bound on the negative log-likelihood of the data. 


DDPMs have demonstrated state-of-the-art performance in generating high-quality samples across various domains, including images, videos, and time-series data.

\section{Methodology}
\begin{figure*}[t]
    \centering
    \includegraphics[width=\linewidth]{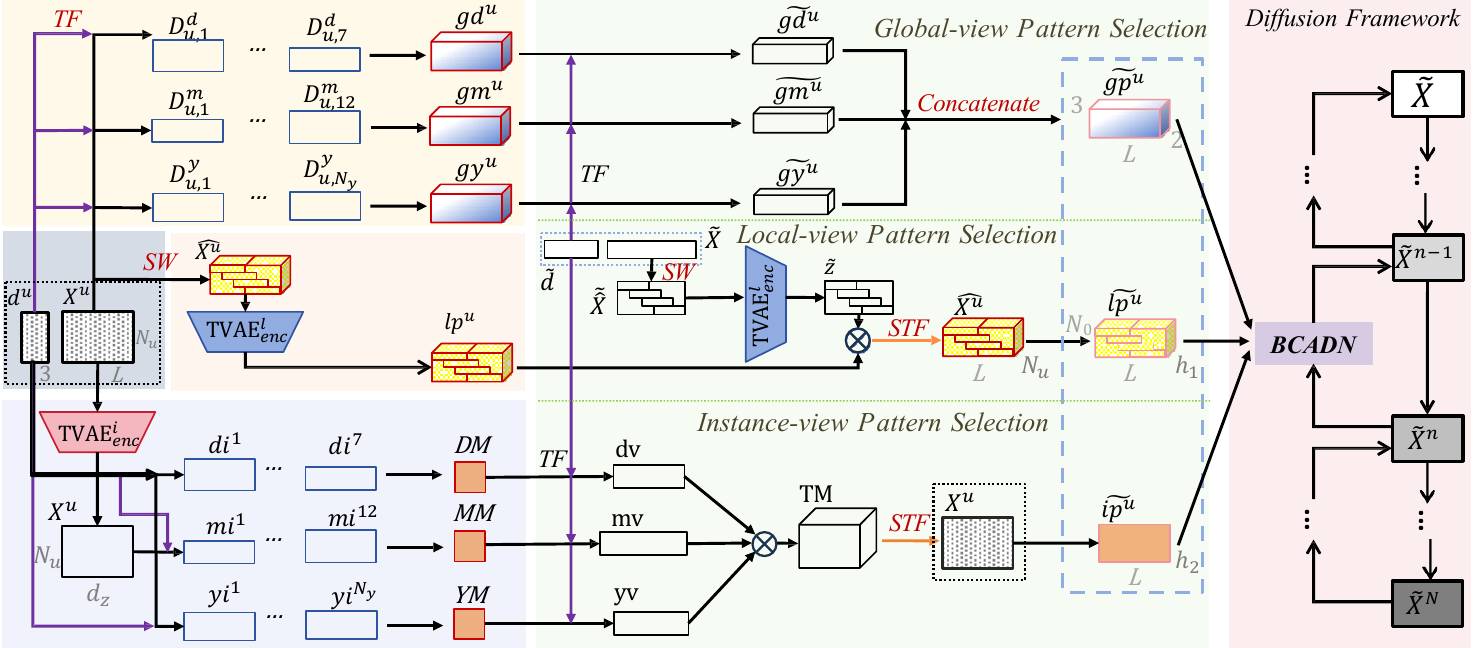}
    \caption{The pipeline of \modelAbvName. The shapes outlined by the red and pink lines represent the entire and selected user behavior patterns, respectively. The black dashed box and blue dashed box represent the historical user-specific utility data and the input of the imputation stage. \textit{TF} refers to data filtering based on time (day, month, year). \textit{SW} represents the splitting of subsequences using sliding windows. \textit{STF} between two tensors $A$ → $B$ denotes the process of selecting entries from $A$ based on the top values in $B$. The detailed architectures of TVAE and BCADN can be found in Figures~\ref{fig:tvae} and ~\ref{fig:denoising}, respectively. Notations in grey denote the dimensional information about the data.
    }
    \label{fig:framework}
\end{figure*}

The overall framework of our proposed \modelAbvName\ is illustrated in Figure~\ref{fig:framework}. \modelAbvName\ comprises two core modules. The first, Multi-view User Behavior Extraction (MUBE), captures global, local, and instance-level behaviors to build a comprehensive user profile. The second, Behavior-aware Conditional Diffusion Model (BCDiff), identifies the most informative features from the profile and leverages them to impute missing data in a computationally efficient manner.
\vspace{-3pt}
\subsection{Multi-view \mOneFullName} \label{sec:extraction}
In this part, we design a multi-view user behavior extraction module \ (\mOneAbvName) to learn comprehensive user behavior from different aspects (e.g., global, local, and instance view) based on long-term and diverse historical utility data records, which will provide rich information for effective data imputation. 
Specifically, for a utility time series $D_u = \{(X_1^u, d_1^u, m_1^u), \cdots, (X_{N_u}^u, d_{N_u}^u, m_{N_u}^u)\}$ collected from user $u$, we target to construct a user profile \(P = (GP, LP, IP)\), where \(GP, LP, IP\) captures the global, local, and instance-view behavior patterns.

\vspace{-3pt}
\subsubsection{Global-view Behavior Pattern Extraction}
Global-view behavior patterns provide prior knowledge for user $u$, offering coarse-grained bounds for each data point in the utility time series and thereby alleviating the incompleteness of historical data records.  
We define three types of global-view patterns for each feature $i$ based on aggregation over the same day of the week, month, and year.
For the daily pattern $gd^u_d$, the mean $\mu_{d,i}$ and variance $\sigma^2_{d,i}$ are calculated over all data points where the day of the week is $d\in [1, 7]$:
\begin{equation}
    gd^u = \{ gd^u_1, \cdots, gd^u_7\}, 
\end{equation}
\begin{equation}
    gd^u_d = \{(\mu_{d,i}^u, {\sigma^u_{d,i}}^2), \cdots, (\mu_{d,L}^u, {\sigma^u_{d,L}}^2)\}, 
\end{equation}
\begin{equation}
    \mu_{d,i}^u = \frac{1}{|D_{u,d}|} \sum_{(X_k^u, d_k^u) \in D_{u,d}} X_{k, i}^u, 
\end{equation}
\begin{equation}
    {\sigma^u_{d,i}}^2 = \frac{1}{|D_{u,d}|} \sum_{(X_k^i, d_k^i) \in D_{u,d}} ( X_{k, i}^u - \mu_{d,i}^u)^2,
\end{equation}
where $D_{u,d}$ is a subset of user $u$'s historical data $D_u$; the day of the week $d_k^i$ is equal to $d$ and $k \in [1, |D_{u,d}|]$; \(X_{k, i}^u\) denotes the $i$-th value in data record \(X_k^u\) and \(L\) represents the length of the utility data.
The monthly patterns \(gm^u_m\) and yearly \(gy^u_y\) patterns are computed in a similar way, where \(m \in [1, 12]\) and \(y\) belongs to the year set \(Y\) of collected data.
These global-view behavior patterns \(GP = (gp^1, gp^2, \cdots, gp^{|U|})\) represent the typical user behavior across different time scales and form a statistical foundation for predicting or imputing values of data points, where $gp^u = [gd^u, gm^u, gy^u]$.

\subsubsection{Local-view Behavior Pattern Extraction}
\label{sss:dp}
\begin{figure}[t]
  \centering
  \includegraphics[width=0.98\linewidth]{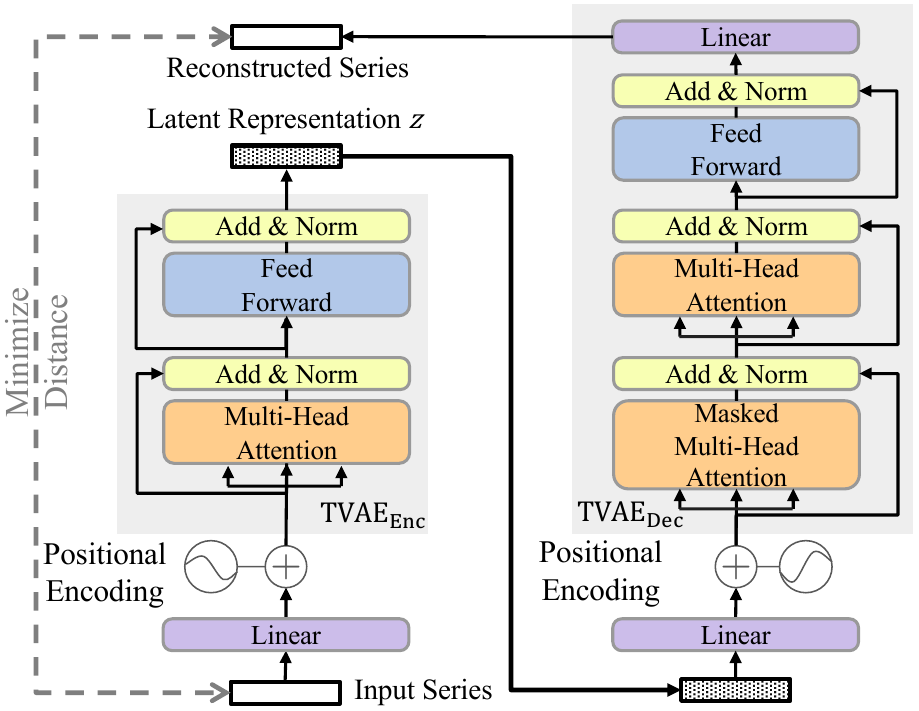}
  \caption{The architecture of the designed TVAE for utility time series embedding and subsequence embedding.}
  \label{fig:tvae}
\end{figure}

Although global-view behavior patterns offer a possible range for each data point, they fail to capture the temporal dependencies between data points in the data record (i.e., a time series), which are important for accurate missing value imputation. 
To address this, we introduce fine-grained local-view behavior patterns, extracted from consecutive data points in $D_u$, that reflect the correlations between these points.
Initially, due to the potential incompleteness of data points, we utilize the global-view patterns \(g^u\) for missing value \(X_{i,j}^u\) filling in each $X_i^u$, where \(m_{i,j}^u = 0\).
Specifically, we obtain pseudo-imputed utility data record $\hat{X_i^u}$ from original data \((X_i^u, d_i^u, m_i^u)\) through:
\begin{equation}
    \hat{X_{i,j}^u} = 
\left\{
\begin{array}{ll}
   gd_{d',j}^u \text{ or } gm_{m',j}^u \text{ or } gy_{y',j}^u & \text{if } m_{i,j}^u = 0 \\
   X_{i,j}^u & \text{if } m_{i,j}^u = 1
\end{array}
\right.
\label{formula:pseudo}
\end{equation}
where \(j\) is the index in the data record, and \(d', m', y'\) are the day, month, and year indices of \(d_i^u\).  
Next, we apply a sliding window of size \(l_w\) and step \(s\) to extract subsequences from each pseudo-imputed utility record \(\hat{X}_i^u\), represented as:
\begin{equation}
    w^u_i = \{w^u_{i,j}\}_{j=1}^{N_w}, \quad
     w^u_{i,j} = \{\hat{X}^u_{i, s_j}, \cdots, \hat{X}^u_{i, e_j}\}
\label{formula:sliding}
\end{equation}

where \(w^u_i\) denotes all subsequences of user \(u\)'s \(i\)-th historical utility data record, and \(w^u_{i,j}\) represents the \(j\)-th subsequence. The starting and ending indices are calculated as \(j_{\mathrm{start}} = s \times (j{-}1)\), \(j_{\mathrm{end}} = j_{\mathrm{start}} + l_w - 1\), and the number of subsequences is given by \(N_w = \left\lfloor \frac{T - l_w}{s} \right\rfloor + 1\). By splitting a full record into multiple subsequences, we extract behavioral segments that capture local dependencies across different positions and offer more fine-grained local references for imputation. We then design a Transformer-based Variational Autoencoder (TVAE, whose architecture is shown in Figure~\ref{fig:tvae}) to learn a unified latent representation \(z^u_{i,j}\) for each subsequence \(w^u_{i,j}\):
\begin{equation}
    z^u_{i,j} = \text{TVAE}_{\text{encoder}}^d(w^u_{i,j}), \quad z^u_{i,j} \in \mathbb{R}^{d_z},
\label{formula:VAE1}
\end{equation}
where \(d_z\) is the dimension of the latent representation, and \(\text{TVAE}_{\text{encoder}}^d\) is the encoder of TVAE for dynamic patterns. 
We train \(\text{TVAE}^d\) on subsequences \(W = \{w^1, w^2, \cdots, w^{|U|} \}\) from all users. 
Finally, we represent dynamic behavioral patterns \(LP = (lp^1, lp^2, \cdots, lp^{|U|})\) by aggregating over subsequence positions in each data record:

\begin{equation} 
    lp^u = \{lp^u_j\}_{j=1}^{N_w}, \quad 
    lp^u_j = \{(w^u_{i,j}, z^u_{i,j})\}_{i=1}^{N_u},
\label{formula:dp}
\end{equation}

\subsubsection{Instance-view Behavior Pattern Extraction}
We design an instance-view behavior pattern extraction module to provide record-level information for imputation, offering a broader context than local-view patterns.  
Specifically, we employ another Transformer-based Variational Autoencoder, denoted as \(\text{TVAE}^i\), to encode the \(i\)-th pseudo-imputed utility record of user \(u\), \(\hat{X}_i^u\), into a shared latent space:  
\(v_i^u = \text{TVAE}_{\text{encoder}}^i(\hat{X}_i^u)\).  
The model \(\text{TVAE}^i\) is trained on all users’ pseudo-imputed records, \(\hat{X} = \{\hat{X}^1, \hat{X}^2, \cdots, \hat{X}^{|U|}\}\).  
To capture instance-view behavior patterns with respect to various time points \(\tilde{d}\) in the target record, we design a multi-dimensional temporal matching mechanism based on three matching matrices:  
\(ip^u = (DM^u, MM^u, YM^u)\), where \(DM^u \in \mathbb{R}^{7 \times 7}\), \(MM^u \in \mathbb{R}^{12 \times 12}\), and \(YM^u \in \mathbb{R}^{N_y \times N_y}\) correspond to day, month, and year dimensions, respectively.  
We first compute the Euclidean distance matrix \(\mathbf{D}^u \in \mathbb{R}^{N_u \times N_u}\), where each element is given by  
\(D_{i,j}^u = \| \mathbf{v}_i^u - \mathbf{v}_j^u \|_2\).  
For each time pair \((p, q)\), we then aggregate the relevant distances to form the three temporal matching matrices \(DM^u\), \(MM^u\), and \(YM^u\), as defined below:

\begin{equation}
    DM^u_{p,q} = \frac{1}{|di^p| \times |di^q|} \sum_{a=1}^{|di^p|} \sum_{b=1}^{|di^q|} Dist^u_{di^p_a, di^q_b},
\end{equation}
\begin{equation}
    MM^u_{p,q} = \frac{1}{|mi^p| \times |mi^q|} \sum_{a=1}^{|mi^p|} \sum_{b=1}^{|mi^q|} Dist^u_{mi^p_a, mi^q_b},
\end{equation}
\begin{equation}
    YM^u_{p,q} = \frac{1}{|yi^p| \times |yi^q|} \sum_{a=1}^{|yi^p|} \sum_{b=1}^{|yi^q|} Dist^u_{yi^p_a, yi^q_b},
\end{equation}
where \(di^p\), \(mi^p\), \(yi^p\) denotes the index set where each index \(di^p_j\), \(mi^p_j\), \(yi^p_j\) satisfies \(dd^u_{di^p_j} = p\), \(md^u_{mi^p_j} = p\), \(yd^u_{yi^p_j} = p\), respectively. 
Finally, each matrix is normalized through the row, indicating the overall historical dependency between each time $p$ and $q$ in three temporal dimensions. 

\subsection{\mTwoFullName}
In this part, we design a \mTwoFullName\ called \mTwoAbvName\ to impute missing values in utility data $\tilde{X}$.
Due to the high-dimensional nature of the constructed user profile $P$, directly applying diffusion models would incur considerable computational cost. To address this, we first introduce a reference selection module that identifies the most informative behavioral patterns from the extracted user profile. We then develop a probabilistic diffusion model equipped with a novel behavior-aware conditional attentional denoising network, which is guided by the selected reference behaviors to impute missing values.

\subsubsection{Reference Selection}

First, we select reference global-view patterns \(\tilde{gp^u} = [\tilde{gd^u},\tilde{gm^u},\tilde{gy^u}] \in \mathbb{R}^{3\times L \times 2}\) from \(gp^u\) by indexing with the three input temporal elements. Specifically, \(\tilde{gd^u} = gd^u_{\tilde{dd}}\), \(\tilde{gm^u} = gm^u_{\tilde{md}}\) , and \(\tilde{gy^u} = gy^u_{\tilde{yd}}\) denote the selected reference global day, month, and year patterns, respectively. Each element \(\tilde{gp^u}_{i,j,0}\) and \(\tilde{gp^u}_{i,j,1}\) represents the mean and variance, respectively.

Next, to obtain reference local-view behavior patterns, we first generate the pseudo-imputed utility data record \(\hat{\tilde{X}}\) and extract unified latent representation pairs \(\{(\tilde{w_1}, \tilde{z_1}), \cdots, (\tilde{w_{N_w}}, \tilde{z_{N_w}})\}\) from \(\tilde{X}\), following Equations~\ref{formula:pseudo}, \ref{formula:sliding}, and \ref{formula:VAE1}. We compute the Euclidean distance between each latent representation \(\tilde{z_i}\) and each historical latent representation in \(lp_i^u\) illustrated in Equation \ref{formula:dp} and obtain a local dynamic set of size $h_2$ \(\tilde{w_i} = \{\tilde{w_{i,1}}, \cdots, \tilde{w_{i,h_2}}\}\) for each location of subsequences, where $i$ denotes the index of subsequence in the data record. 
The entire selective reference local-view behavior patterns are represented as \(\tilde{lp^u} = \{ \tilde{w_1}, \cdots, \tilde{w_{N_w}}\}\).

In terms of the selection of instance-view behavior patterns, we initially obtain three weighted vectors \(dv \in \mathbb{R}^7\), \(mv \in \mathbb{R}^7\), \(yv \in \mathbb{R}^{N_y}\) from corresponding rows, where \(dv =DM^u_{\tilde{dd},:} \), \(mv =MM^u_{\tilde{dd},:} \), \(yv =YM^u_{\tilde{yd},:}\).
A temporal weight matrix \(TM\) is computed through the outer product of \(dv\), \(mv\), \(yv\) as follows:
\begin{equation}
 \mathbf{TM}_{i,j,k} = \mathbf{dv}_i \cdot \mathbf{mv}_j \cdot \mathbf{yv}_k, \quad
 \mathbf{TM} \in \mathbb{R}^{7 \times 12 \times N_y},
\end{equation}

We then select \(h_2\) samples \(\tilde{ip}^u = \left\{ \left(\tilde{X}_i^u, \tilde{d}_i^u, \tilde{m}_i^u \right) \right\}_{i=1}^{h_2}\), whose dates correspond to the top \(h_2\) weights in the temporal matching matrix \(TM\), from the user's historical utility data \(D_u\) as reference instance-view behavior patterns.

In summary, three types of selected behavior patterns \(\tilde{gp^u} \in \mathbb{R}^{3\times L \times2}\), \(\tilde{lp^u} \in \mathbb{R}^{N_w\times h_2 \times l_w}\), \(\tilde{ip^u} \in \mathbb{R}^{h_2 \times L}\) serve as the references that steer the imputation process of the target utility data records.

\subsubsection{Conditional Probabilistic Diffusion Imputation}
\begin{figure}[t]
  \centering
  \includegraphics[width=0.99\linewidth]{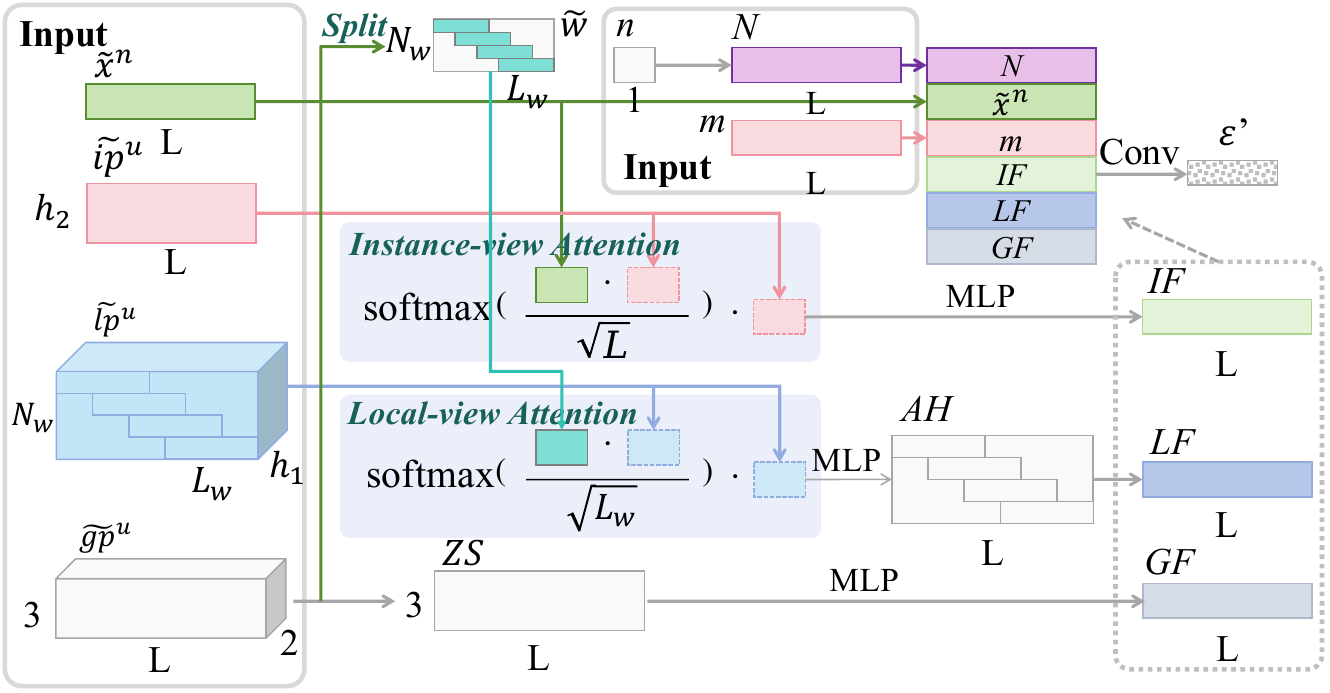}
  \caption{The architecture of the proposed behavior-aware conditional attentional denoising network (BCADN).}
  \label{fig:denoising}
\end{figure}

We propose a steered conditional probabilistic diffusion model armed with all sorts of guidance specifically chosen for $\tilde{X}$ to impute missing values.
Within the diffusion model, we design a novel behavior-aware conditional attentional denoising network (BCADN) that fully extracts the multi-types of correlations among missing values, observed values, and chosen references.
The architecture of BCADN is shown in Figure~\ref{fig:denoising}. 
Given the noisy utility data record $\tilde{X}^n$, at diffusion step $n$, observation mask $\tilde{m}$ and selected patterns \(\tilde{gp^u}\), \(\tilde{lp^u}\), \(\tilde{ip^u}\), the noise prediction network \(\epsilon_\theta\) predicts the added noise \(\epsilon' = \epsilon_\theta(\tilde{gp^u}, \tilde{lp^u}, \tilde{ip^u}, n, \tilde{m}, \tilde{X}^n)\).
We compute Z-scores \(ZS\) using selected global patterns \(\tilde{gp^u}\) as the reference distribution to measure the deviation of observed values. Based on these scores, we derive an intermediate global pattern-steered feature \(GF\) that captures the overall deviation across the three temporal dimensions, which are formalized as:
\begin{equation}
    ZS_{i,j} = \frac{X_j - \tilde{gp^u}_{i,j,0}}{\tilde{gp^u}_{i,j,1}},\; GF = \text{MLP}(\text{Norm}(ZS)), 
\end{equation}
where \(ZS \in \mathbb{R}^{3 \times L}\), \(GF \in \mathbb{R}^{L}\), MLP denotes the multilayer perceptron, and Norm represents the normalization operation along the row dimension.
To capture the dependency between subsequences of $\tilde{X}$ or the entire $\tilde{X}$ and either local-view or instance-view patterns, we design Attention layers that capture the correlation between adjacent or distant data points as follows:

\begin{align}
    H^l &= \big[\text{Attn}(\tilde{w}_j, \tilde{lp}^u_j, \tilde{lp}^u_j)\big]_{j=1}^{N_w}, \\
    H^i &= \text{Attn}(\tilde{X}, \tilde{ip}^u, \tilde{ip}^u), \;
    H^l \in \mathbb{R}^{N_w \times l_w},\; H^i \in \mathbb{R}^{L}
\end{align}
where \(\text{Attn}(Q, K, V) = \text{softmax}(\frac{Q\cdot K}{\sqrt{l}})\cdot V\) represents the attention layer and \(l\) is equal to \(l_w\) and \(L\) for \(H^l\) and \(H^i\) respectively. 
Considering the different locations of row vectors in \(H^l\) representing the subsequences in the time series, we align them at corresponding positions as follows:
\begin{equation}
    AH_{i,j} = 
\left\{
\begin{array}{ll}
   H^d_{i, j-i_{start}} & \text{if} \quad 0\leq j-i_{start}<l_w\\
   0 & \text{otherwise}
\end{array}
\right.
\end{equation}
where \(AH^d \in \mathbb{R}^{N_w \times L}\) and \(i_{start} = s \times (i - 1)\) represents the starting index of $i$-th subsequence. 
The intermediate local-view and instance-view pattern-steered feature \(LF = \text{MLP}(AH^T) \in \mathbb{R}^L\) and \(IF = \text{MLP}(H^i) \in \mathbb{R}^L\) are derived through MLP, which naturally incorporates the positional information without the need to introduce positional embedding to lower the computational complexity.
Finally, we fuse three intermediate features, diffusion step \(N = Pos(n) \in L\) embedded with positional encoding and observation mask $\tilde{m}$ with noisy data record \(\tilde{X}^n\) from previous step and obtain the predicted noise \(\epsilon' = \text{1DConv}(Concat(\tilde{X}^n, N, \tilde{m}, GF, LF, IF)) \in \mathbb{R}^L\).
We train \(\epsilon_\theta\) by minimizing the divergence between the added ground-truth noise \(\epsilon\) and predicted noise \(\epsilon'\).

\section{Evaluation}

\subsection{Evaluation Settings}
\subsubsection{Dataset}

We evaluate our \modelAbvName\ using three different types of utility datasets (gas, water, electricity), which are collected from over 120K users. The water and gas datasets are sparser (containing more zeros) compared to the electricity data.
Each data record represents a user’s utility usage time series for a day. We preprocess the three datasets by filtering out users with fewer than 200 records.
Utility data records for each user are randomly split into training, validation, and test sets with a 7:1:2 ratio, ensuring a balanced distribution across users. 
A data statistics of the three datasets is summarized in Table~\ref{tab:dataset}.

\begin{table}[!h]
\centering \small
\caption{Statistics of the Three Utility Datasets.}
\begin{tabular}{c|lll}
\hline
 & \textbf{Electricity} & \textbf{Water} & \textbf{Gas} \\ \hline
Length of Each Series       & 48 & 24 & 24 \\ \hline
Sampling Interval       & 30 mins & 60 mins & 60 mins \\ \hline
Number of Users      & 60,907 & 90,704 &  31,683\\ \hline
Mean \# Series per User   & 701 & 317 & 709 \\ \hline
Data Duration     & 2018-2019 & 2019 & 2019-2020 \\ \hline
\end{tabular}
\label{tab:dataset}
\end{table}

\begin{table*}[ht]
\centering \small
\caption{Evaluation results on three different datasets with a missing rate of 20\%.} 
\label{tab:mae&mse}
\begin{tabular}{lcccccccccccc}
\toprule
& \multicolumn{4}{c}{\textbf{Electricity}} & \multicolumn{4}{c}{\textbf{Gas}} & \multicolumn{4}{c}{\textbf{Water}} \\
\cmidrule(lr){2-5} \cmidrule(lr){6-9} \cmidrule(lr){10-13}
\textbf{Method} & \multicolumn{2}{c}{Point Missing} & \multicolumn{2}{c}{Block Missing} & \multicolumn{2}{c}{Point Missing} & \multicolumn{2}{c}{Block Missing} & \multicolumn{2}{c}{Point Missing} & \multicolumn{2}{c}{Block Missing} \\
\cmidrule(lr){2-3} \cmidrule(lr){4-5} \cmidrule(lr){6-7} \cmidrule(lr){8-9} \cmidrule(lr){10-11} \cmidrule(lr){12-13}
& MAE & MSE & MAE & MSE & MAE & MSE & MAE & MSE & MAE & MSE & MAE & MSE \\ 
\midrule
G-Mean      &0.5021 &0.4530 &0.5062 &0.4781 &0.9963 &1.6034 &0.9812 &1.4233 &1.0410 &1.1196 &1.0476 &1.1347 \\
U-Mean      &0.3773 &0.3045 &0.3813 &0.3176 &0.5946 &5.0742 &0.5350 &4.3112 &0.6014 &0.6076 &0.6040 &0.6080 \\
LINEAR      &0.2130 &0.1255 &0.2706 &0.2115 &0.1223 &0.2610 &0.1349 &0.2874 &0.0316 &0.4920 &0.0317 &0.0286 \\
KF          &0.3286 &0.5099 &0.5020 &0.7879 &0.1899 &1.1888 &0.2250 &0.8421 &0.0405 &0.0351 &0.0460 &0.0523 \\
SAITS       &0.2170 &0.0141 &0.2019 &0.0438 &0.1295 &0.1871 &0.1267 &0.1970 &0.3041 &0.0941 &0.1740 &0.0320 \\
GP-VAE      &0.0382 &0.0031 &0.0383 &0.0034 &\textbf{0.0187} &\textbf{0.0025} &0.0263 &\textbf{0.0020} &0.0271 &0.0075 &0.0432 &\underline{0.0070} \\
CSDI        &0.0293 &\underline{0.0025} &\underline{0.0301} &\underline{0.0031} &0.0267 &0.0097 &0.0214 &0.0078 &\textbf{0.0196} &0.0071 &\underline{0.0285} &0.0112 \\
ImputeFormer&0.1347 &0.0952 &0.3328 &0.1252 &0.1465 &0.0549 &0.2537 &0.0893 &0.0472 &0.0086 &0.2323 &0.1514 \\ 
PriSTI      &0.0288 &0.0031 &0.0326 &0.0034 &\underline{0.0193} &\underline{0.0036} &\underline{0.0213} &0.0089 &\underline{0.0204} &\underline{0.0069} &0.0301 &0.0091 \\ 
TimeMixer++ &\textbf{0.0199} &0.0031  &0.1561 &0.1368 &0.1134 &0.0187 &0.1282 &0.0646 &0.0751 &0.0104 &0.0923 &0.0532 \\
\midrule
\modelAbvName &\underline{0.0276} &\textbf{0.0024} &\textbf{0.0293} &\textbf{0.0028} &0.0213 &0.0085 &\textbf{0.0198} &\underline{0.0069} &0.0251 &\textbf{0.0054} &\textbf{0.0202} &\textbf{0.0065} \\ 
\bottomrule
\end{tabular}
\end{table*}



\subsubsection{Baselines}
We compare our \modelAbvName\ with ten representative baseline methods covering statistical, deterministic, and generative approaches:
\begin{itemize}
    \item \textbf{G-MEAN}~\citep{amiri2016missing}: Imputes missing values using the global historical average across all users.
    \item \textbf{U-MEAN}~\citep{amiri2016missing}: Imputes missing values using the target user's historical average.
    \item \textbf{LINEAR}~\citep{huang2021missing}: Performs linear interpolation to fill missing values.
    \item \textbf{KF}~\citep{bakibillah2022robust}: Uses Kalman Filtering to estimate and impute missing entries.
    \item \textbf{SAITS}~\citep{du2023saits}: A deterministic imputation model based on self-attention mechanisms for time series.
    \item \textbf{GP-VAE}~\citep{fortuin2020gp}: A probabilistic model combining variational autoencoders and Gaussian Processes to capture temporal dependencies.
    \item \textbf{CSDI}~\citep{tashiro2021csdi}: A conditional diffusion model utilizing Transformer-based denoising for time series imputation.
    \item \textbf{PriSTI}~\citep{liu2023pristi}: Enhances CSDI by incorporating spatiotemporal attention to better model spatial and temporal correlations.
    \item \textbf{ImputeFormer}~\citep{nie2024imputeformer}: A low-rank Transformer that balances model expressiveness and inductive bias for spatiotemporal imputation tasks.
    \item \textbf{TimeMixer++}~\citep{wang2024timemixer++}: A general-purpose time series modeling architecture designed for tasks such as forecasting, classification, and imputation.
\end{itemize}

\subsubsection{Metrics}
We use three metrics to evaluate the overall imputation performance: Mean Absolute Error (MAE), Mean Squared Error (MSE), and Continuous Ranked Probability Score (CRPS).
MAE quantifies the average magnitude of the errors between the imputed values and the ground truth data, defined as $\text{MAE} = \frac{1}{n} \sum_{i=1}^n | \hat{x}_i - x_i |$, where $x_i$ is the true value, $\hat{x}_i$ is the imputed value, and $n$ is the total number of instances.
MSE evaluates the average squared difference between the imputed and true values, $\text{MSE} = \frac{1}{n} \sum_{i=1}^n (\hat{x}_i - x_i)^2$; unlike MAE, it penalizes larger deviations more heavily, making it sensitive to outliers.
CRPS measures the probabilistic accuracy of imputed values when the imputation method provides an estimated probability distribution rather than a single point estimate, thus capturing uncertainty as well. Given a cumulative distribution function (CDF) $F$ and true value $x$, it is defined as 
\begin{equation}
\operatorname{CRPS}(F,x)
=
\int_{-\infty}^{\infty}
\left[
F(y)-\mathbb{I}\{y \geq x\}
\right]^2
\,\mathrm{d}y,
\end{equation}
where $\mathbb{I}(y \geq x)$ is an indicator function and a lower value indicates better calibration and sharpness of the predicted distribution.
Among them, MAE and MSE are applied to all methods, while CRPS i s only used for probabilistic models such as GP-VAE, CSDI, PriSTI, and our \modelAbvName.

\subsubsection{Implementation Details}
\label{app:details}
We implemented the \modelAbvName\ framework in PyTorch and conducted all experiments on an NVIDIA A100 GPU. The model is trained using a two-stage process. 
In the first stage, we train the TVAE models for behavior pattern extraction. The utility consumption data is structured into tensors of shape $\mathbb{R}^{144 \times 48}$, representing three days of readings taken at 30-minute intervals. Local patterns are subsequently extracted using a sliding window with a length of 24 and a stride of 12. These TVAE models, configured with a latent dimension of $d_z = 64$, 4 attention heads, and 2 Transformer layers, are trained for 100 epochs using the Adam optimizer with a learning rate of $\alpha = 0.001$ and a weight decay of $\lambda = 10^{-6}$.
In the second stage, we train the diffusion model. This stage runs for 200 epochs using $T = 500$ diffusion timesteps and a linear noise schedule with $\beta_1$ starting at $10^{-4}$ and $\beta_T$ ending at $0.5$. The learning rate is decayed at the 75\% and 90\% training milestones. For all training procedures, we used a batch size of 16 and performed validation every 20 epochs.

For the baselines, parameters were configured to ensure a robust comparison. 
The statistical methods G-MEAN \citep{amiri2016missing}, U-MEAN \citep{amiri2016missing}, and LINEAR \citep{huang2021missing} are hyperparameter-free. For the KF \citep{bakibillah2022robust} model, the process and measurement noise covariance matrices were initialized as identity matrices. The deep learning models were all trained using the Adam optimizer with a learning rate of $1 \times 10^{-3}$ and a batch size of 32. Specifically, SAITS \citep{du2023saits} was configured with 2 attention layers, 4 heads, and a hidden dimension of 128. GP-VAE \citep{fortuin2020gp} used a latent dimension of 64 and a Radial Basis Function (RBF) kernel for its Gaussian Process component. The diffusion models, CSDI \citep{tashiro2021csdi} and PriSTI \citep{liu2023pristi}, were implemented with $T=100$ diffusion steps and a 4-layer Transformer backbone. Lastly, both ImputeFormer \citep{nie2024imputeformer} and TimeMixer++ \citep{wang2024timemixer++} were structured with 4 layers, a model dimension of 128, and 4 attention heads.



\begin{table}[t]
\small
\caption{The results of CRPS on three different datasets.}
\label{tab:crps}
\setlength{\tabcolsep}{2.7pt}
\begin{tabular}{lcccccc} 
\toprule
\textbf{Method} & \multicolumn{2}{c}{\textbf{Electricity}} & \multicolumn{2}{c}{\textbf{Gas}} & \multicolumn{2}{c}{\textbf{Water}}\\
\cmidrule(lr){2-3} \cmidrule(lr){4-5} \cmidrule(lr){6-7}
\textbf{Missing} & Point & Block  & Point  & Block  & Point  & Block  \\ 
\midrule
GP-VAE      &0.0292 &0.0414 &0.0188 &0.0192 &0.0312 &0.0331   \\
CSDI         &0.0233  &\underline{0.0377} &0.0169 &0.0174 &0.0298 &0.0317   \\
PriSTI     &0.0225 &0.0389 &\underline{0.0135} &\textbf{0.0147} &\underline{0.0257} &\textbf{0.0263}    \\ 
ImputeFormer        &0.1347 &0.3328 & 0.1465&0.2537 &0.472  &0.2323   \\
TimeMixer++        &\underline{0.0199} &0.1561 & 0.1134 &0.1282 &0.0751 &0.0923 \\ \midrule
\modelAbvName &\textbf{0.0168} &\textbf{0.0359} &\textbf{0.0113} &\underline{0.0168} &\textbf{0.0239} &\underline{0.0281} \\
\bottomrule
\end{tabular}
\end{table}

\subsection{Overall Performance}
We first present the imputation results of different methods on the three datasets with 20\% missing data (Tables \ref{tab:mae&mse} and \ref{tab:crps}). To reflect real-world scenarios, we consider two missing patterns: point missing, where individual values are randomly removed, and block missing, where several values are removed in sequence, starting from a random point.

Overall, traditional statistical methods do not perform well on any of the datasets, even with user-specific information. This suggests that user behavior in utility usage is often irregular and hard to predict, especially in the gas and water datasets, which are sparser. In contrast, neural network-based methods perform better by capturing patterns in the data. Among all models, those that generate data rather than predict fixed values tend to produce more stable and accurate results. On the electricity dataset, our method \modelAbvName\ achieves the best performance in both point and block missing cases. 
For example, it achieves an MAE of 0.0293 and an MSE of 0.0024, which are 2.32\% and 4.0\% better than the next-best model in the block and point missing setting, respectively. 
It also improves CRPS by 15.57\%, 7.71\% over the second-best one in the same settings.

Electricity usage records are relatively dense and consistent, which helps \mOneAbvName\ identify and use meaningful patterns for filling in missing values. On the other hand, gas and water data are sparser and harder to learn from. Even so, \mOneAbvName\ still ranks in the top two across all settings. For gas data, GP-VAE works better on point missing cases, while \modelAbvName\ performs better on block missing. On water data, \modelAbvName\ shows strong advantages. For instance, it reduces block missing MAE by 29.1\% compared to CSDI, and improves point missing CRPS by 7\% over PriSTI, showing its ability to handle more irregular and incomplete data.

\begin{table}[h]
\centering
\small
\caption{Sensitivity analysis under different missing ratios (point missing).}
\label{tab:sensitivity_small}
\setlength{\tabcolsep}{2.5pt}
\begin{tabular}{lcccccc}
\toprule
\multirow{2}{*}{\textbf{Method}} 
& \multicolumn{2}{c}{\textbf{20\%}} 
& \multicolumn{2}{c}{\textbf{35\%}} 
& \multicolumn{2}{c}{\textbf{50\%}} \\
\cmidrule(lr){2-3} \cmidrule(lr){4-5} \cmidrule(lr){6-7}
& MAE & MSE & MAE & MSE & MAE & MSE \\
\midrule
GP-VAE         
& 0.0382 & 0.0031 
& 0.0523 & 0.0049 
& 0.0522 & 0.0068 \\

CSDI           
& 0.0293 & \underline{0.0025} 
& 0.0514 & 0.0055 
& 0.0519 & 0.0059 \\

PriSTI         
& 0.0288 & 0.0031 
& \underline{0.0402} & \textbf{0.0043} 
& \underline{0.0472} & \underline{0.0052} \\

ImputeFormer   
& 0.1347 & 0.0952 
& 0.1480 & 0.1025 
& 0.1567 & 0.1236 \\

TimeMixer++    
& \textbf{0.0199} & 0.0031 
& 0.0983 & 0.0565 
& 0.1804 & 0.1583 \\
\midrule
\modelAbvName
& \underline{0.0276} & \textbf{0.0024} 
& \textbf{0.0387} & \underline{0.0049} 
& \textbf{0.0396} & \textbf{0.0050} \\
\bottomrule
\end{tabular}
\end{table}

\subsection{Sensitivity Analysis}
To address the diverse types and rates of missing data in real-world scenarios, we evaluate the robustness of each method under varying missing proportions. Due to the sparsity of the water and gas datasets, we focus our analysis on the electricity dataset. As shown in Table~\ref{tab:sensitivity_small}, we report MAE and MSE under point missing settings at different rates (20\%, 35\%, 50\%). Our method, \modelAbvName, consistently ranks first or second across all conditions, outperforming other baselines overall. For example, with a low missing rate of 20\%, aided by the user profile extracted by \mOneAbvName, \mTwoAbvName\ selects more informative local- and instance-view references to guide accurate imputations. Even at a 50\% missing rate, \modelAbvName\ remains robust, achieving an MAE of 0.0396, which is better than PriSTI (0.0472) and CSDI (0.0519). It also achieves the lowest MSE of 0.0050. These results confirm \modelAbvName’s robustness under severe missing conditions. 

To address various missing data types and rates encountered in real-world scenarios, we evaluate the robustness of different methods by assessing their performance across different missing types and imputation rates.
Due to the sparsity of the water and gas datasets, we focus on the electricity dataset in this part. As shown in Tab. \ref{tab:s_mae&mse} and Tab. \ref{tab:s_crps},
at low missing rates (20\%), \modelAbvName\ achieves superior performance across metrics, with the best MAE and MSE in both point and block missing scenarios, improving 4.2\% in point missing MAE compared to PriSTI.
To handle a small portion of data points missing (20\%), given the extracted user profile by \mOneAbvName, \mTwoAbvName\ can select more instructive local-view and instance-view reference that steers the imputation.
With the missing rate increases, the selection of local-view references may be impacted, as the matching subsequences in the data records to be imputed are filled with coarse-grained priors.
At 35\% missing rate, \modelAbvName\ maintains strong performance with the best MAE in both point missing (0.0387, 3.7\% better than PriSTI) and block missing (0.0346, 10.8\% better than PriSTI). Even at a 50\% missing rate, \modelAbvName\ demonstrates remarkable resilience, achieving the best point missing data imputation performance (MAE: 0.0396, improving 16.1\% over PriSTI) and remaining competitive in block missing scenarios.
Despite this, with the aid of global-view and instance-view behavior selected solely based on temporal information, \modelAbvName\ still outperforms other baselines in most cases.
Similarly, Tab. \ref{tab:s_crps} shows the comparison of CRPS across different missing rates. From this table, we found that \modelAbvName\ demonstrates consistent strong performance. At a 20\% missing rate, \modelAbvName\ achieves the best point missing imputation CRPS of 0.0210 (6.7\% improvement over PriSTI). With the missing rate increasing to 35\%, \modelAbvName\ maintains its advantage with the best CRPS in both point (0.0374) and block (0.0386) scenarios. Even at a 50\% missing rate, \modelAbvName\ still shows strong performance with competitive CRPS values, particularly in block missing (0.0479), which is 7.7\% better than PriSTI. These results further validate \modelAbvName's robust performance across different evaluation metrics and missing scenarios.

\begin{table}[h]
\small
\caption{Ablation study of different user behavior views. The abbreviations 'glb', 'loc', and 'ins' refer to the global-view, local-view, and instance-view behavior patterns, respectively.}
\vspace{-5pt}
\label{tab:ablation_MAE}
\setlength{\tabcolsep}{4pt}
\begin{tabular}{lcccccc}
\toprule
\textbf{Method} & \multicolumn{2}{c}{\textbf{20\%}} & \multicolumn{2}{c}{\textbf{35\%}} & \multicolumn{2}{c}{\textbf{50\%}}\\
\cmidrule(lr){2-3} \cmidrule(lr){4-5} \cmidrule(lr){6-7}
 & MAE & MSE  & MAE  & MSE  & MAE  & MSE  \\ 
\midrule
glb + loc      & 0.0280 &  0.0027& 0.0390 & 0.0055 &  0.0416 & 0.0054  \\
glb + ins      & 0.0291 &  0.0035& 0.0401 & 0.0072 &  0.0449  &  0.0060 \\
loc + ins      & 0.0293 &  0.0032 & 0.0395 &  0.0067 &  0.0421  &  0.0053 \\ \midrule
\modelAbvName &\textbf{0.0276} &\textbf{0.0024} &\textbf{0.0387} &\textbf{0.0049} &\textbf{0.0396} &\textbf{0.0050}   \\
\bottomrule
\end{tabular}
\vspace{-10pt}
\end{table}
\subsection{Ablation Study}

To further assess the effectiveness of incorporating three different views of user behaviors in our proposed \modelAbvName, we conduct ablation studies on the electricity dataset, comparing the full model against variants that leverage only two types of extracted behaviors. As shown in Table~\ref{tab:ablation_MAE}, our full \modelAbvName\ model consistently achieves the best performance across all missing rates, demonstrating the strong complementarity of the three views. Notably, removing the local-view behavior (the glb + ins variant) leads to the most significant performance drop, particularly at the severe 50\% missing rate, which indicates that fine-grained, short-term temporal dependencies are crucial for guiding imputation when data becomes highly incomplete. Conversely, the glb + loc variant exhibits the most stable performance among the ablated pairs, suggesting that combining short-term local dependencies with long-term global statistical boundaries provides a highly robust foundation. Ultimately, while eliminating any single behavior pattern leads to a performance decline compared to the full model, these ablated versions still maintain competitive performance against most standard baselines, validating that a comprehensive, multi-view approach is essential for accurate utility data imputation.


\section{Related Work}

\subsection{Time Series Data Imputation}
Time series data imputation fills in missing values in sequential data to ensure completeness for analysis and applications, and it has been widely studied in various fields such as finance, healthcare, and transportation. 
In the early years, traditional methods such as linear interpolation (LinITP)\citep{huang2021missing} and Kalman filtering (KF)\citep{bakibillah2022robust} relied heavily on statistical prediction, but struggled to capture the complex temporal patterns inherent in real-world data.
Recent advances in deep learning have led to rapid progress in time series imputation, which can be broadly divided into predictive and generative approaches. Predictive methods estimate missing values by leveraging temporal context from both past and future observations, typically using reconstruction-based models such as RNNs \cite{cao2018brits}, CNNs \cite{wu2023timesnet}, GNNs \cite{cini2022filling}, or attention mechanisms \cite{shan2023nrtsi}. While effective in capturing local patterns, these models often struggle with scalability and modeling global structures in long sequences.

In this work, we focus on generative methods, as they can produce diverse outputs for missing observations and are better suited to capturing overall data distributions. Among these, VAE-based \cite{kim2023probabilistic} and GAN-based \cite{liu2019naomi,miao2021generative} methods rely on direct sampling from the latent space, which limits their generalization. Recently, diffusion models have demonstrated strong capabilities in modeling complex data through multi-step noise injection. CSDI \cite{tashiro2021csdi} first adopts conditional training using partial observations, 
SSSD \cite{alcaraz2022diffusion} introduces structured state space models, 
and PriSTI \cite{liu2023pristi} incorporates spatiotemporal dependencies. However, these methods do not explicitly model user-specific behaviors. In our work, we address this limitation by incorporating fine-grained user characteristics from historical data as prior knowledge. This approach improves both model performance and real-world applicability.

\subsection{Utility Data Mining}
High-quality utility time series data is essential for real-world applications such as billing, resource allocation, and system optimization. Many data mining tasks, including demand forecasting and anomaly detection, rely on these data to uncover meaningful patterns. In the water systems, DTW-based clustering reveals correlations between consumption behavior and socioeconomic status \cite{steffelbauer2021dynamic}. Smart meter data in gas systems enables pipeline topology inference and leakage detection \cite{matalkah2020smart}. In electricity systems, fine-grained records support load profiling, demand prediction, and operational planning \cite{wang2018review}, and recent studies further advance accurate and reliable user-level energy usage prediction \cite{yu2026trustenergy, yu2026energymamba}.
Utility mining techniques have also been applied in other domains, such as wastewater system analysis \cite{cipolla2014heat} and heating system optimization \cite{calikus2019data}.
These diverse downstream applications fundamentally rely on realistic and high-quality utility data. Therefore, our work lays a vital foundation for a broad range of utility data mining tasks.

\section{Conclusion}
In this paper, we propose \modelAbvName, a novel behavior-aware utility data imputation framework that effectively incorporates comprehensive user behavior to steer accurate missing value imputation. 
\modelAbvName\ consists of two core modules: 
(i) a multi-view user behavior extraction module to learn comprehensive user behavior from different aspects (e.g., global, local, and instance view), and (ii) 
a behavior-aware conditional diffusion model consisting of a reference selection module and a conditional
attentional denoising network to impute utility data in a computationally efficient way.
We implement and evaluate \modelAbvName\ by collaborating with a utility provider, and extensive experimental results demonstrate that \modelAbvName\ outperforms state-of-the-art baselines across three datasets, e.g., it improves 7.04\% and 29.1\% on the electricity and water usage datasets for block missingness imputation, showcasing the effectiveness of utilizing user behavior as guidance for imputation and highlighting its ability to address real-world challenges of missingness in utility data. 
It has great potential to impute missing values in utility data to improve utility services and business intelligence.
Beyond utility time series imputation, behavioral patterns extracted from existing data hold significant potential to guide imputation across various types of temporal or spatio-temporal time series, offering a promising avenue for future exploration.

\section*{Acknowledgment}
We thank all the reviewers for their insightful feedback and comments, which helped us improve this paper. This work is partially supported by the Florida State University (FSU) Startup Fund, FSU Summer Research Support (SRS) Award Program, and
FSU-AWS Research Acceleration Fund.
\bibliographystyle{aaai2026}
\bibliography{main}

\appendix

\appendix
\section{Appendix}

\subsection{Motivation and Data Analysis}
\label{App:motivation}
\begin{figure*}[]
\centering
\begin{subfigure}[b]{0.33\linewidth}  
    \includegraphics[width=\linewidth]{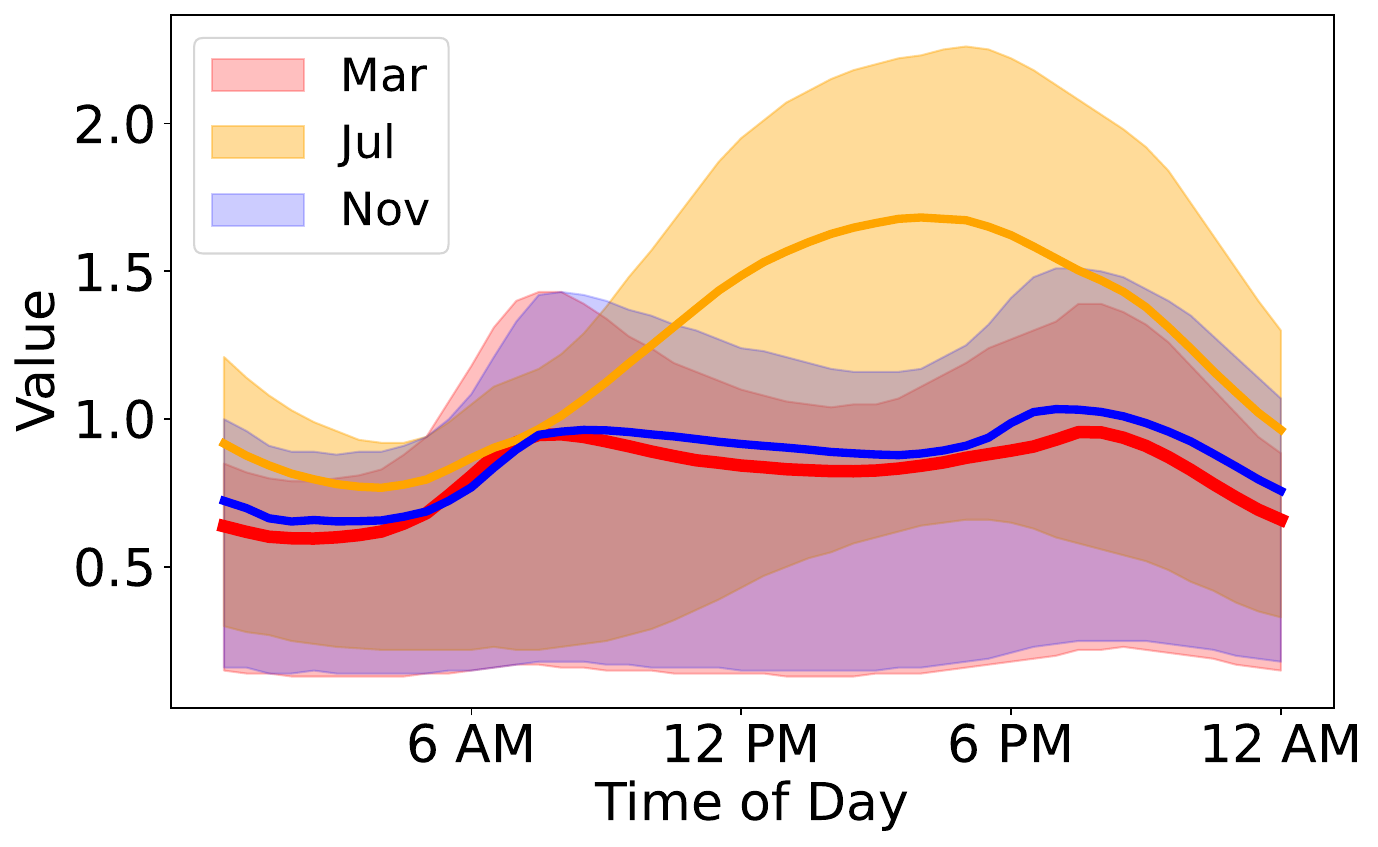}
    \caption{Electricity}
    \label{fig:sub1}
\end{subfigure}
\hfill
\begin{subfigure}[b]{0.33\linewidth}
    \includegraphics[width=\linewidth]{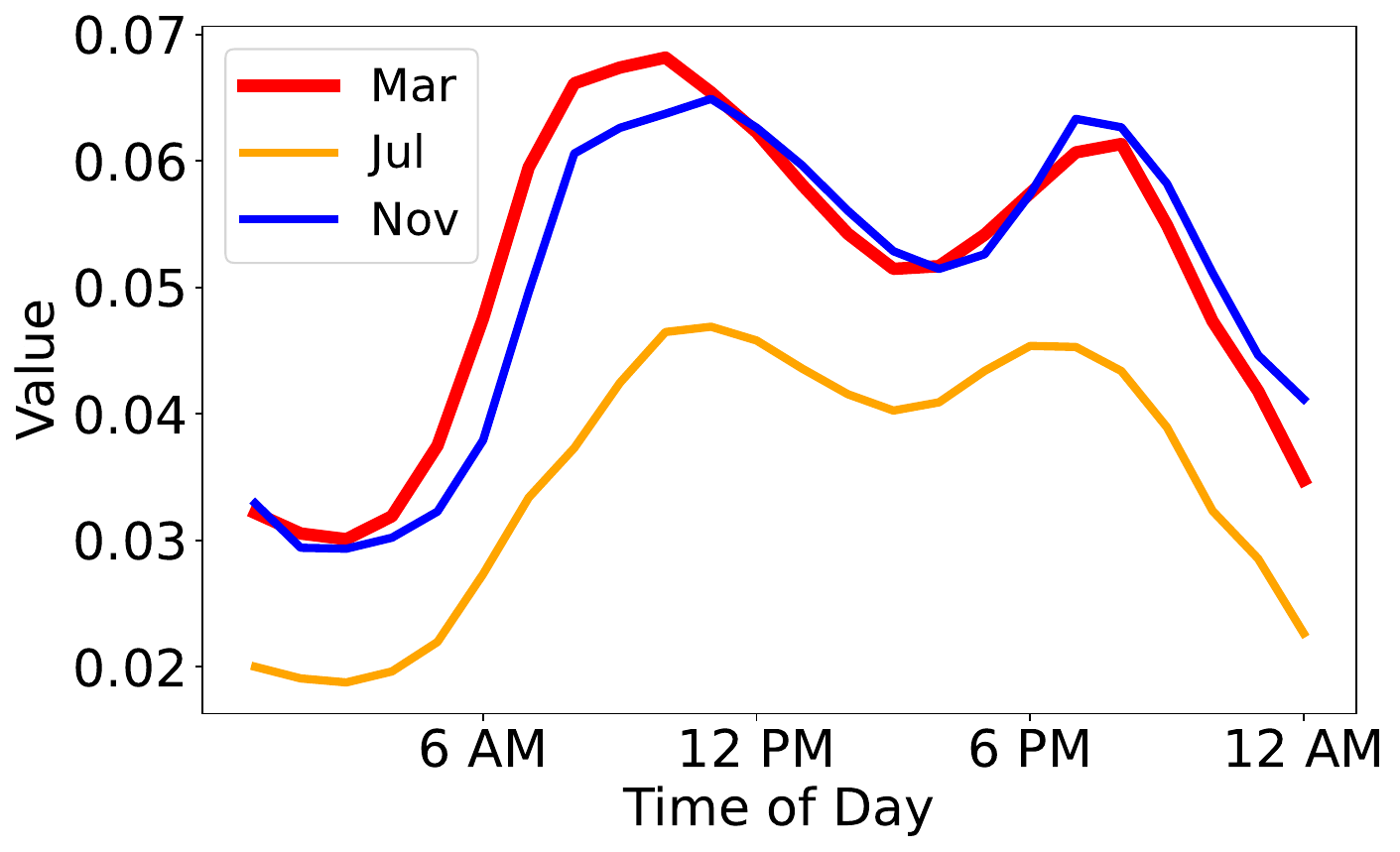}
    \caption{Gas}
    \label{fig:sub2}
\end{subfigure}
\hfill
\begin{subfigure}[b]{0.33\linewidth}
    \includegraphics[width=\linewidth]{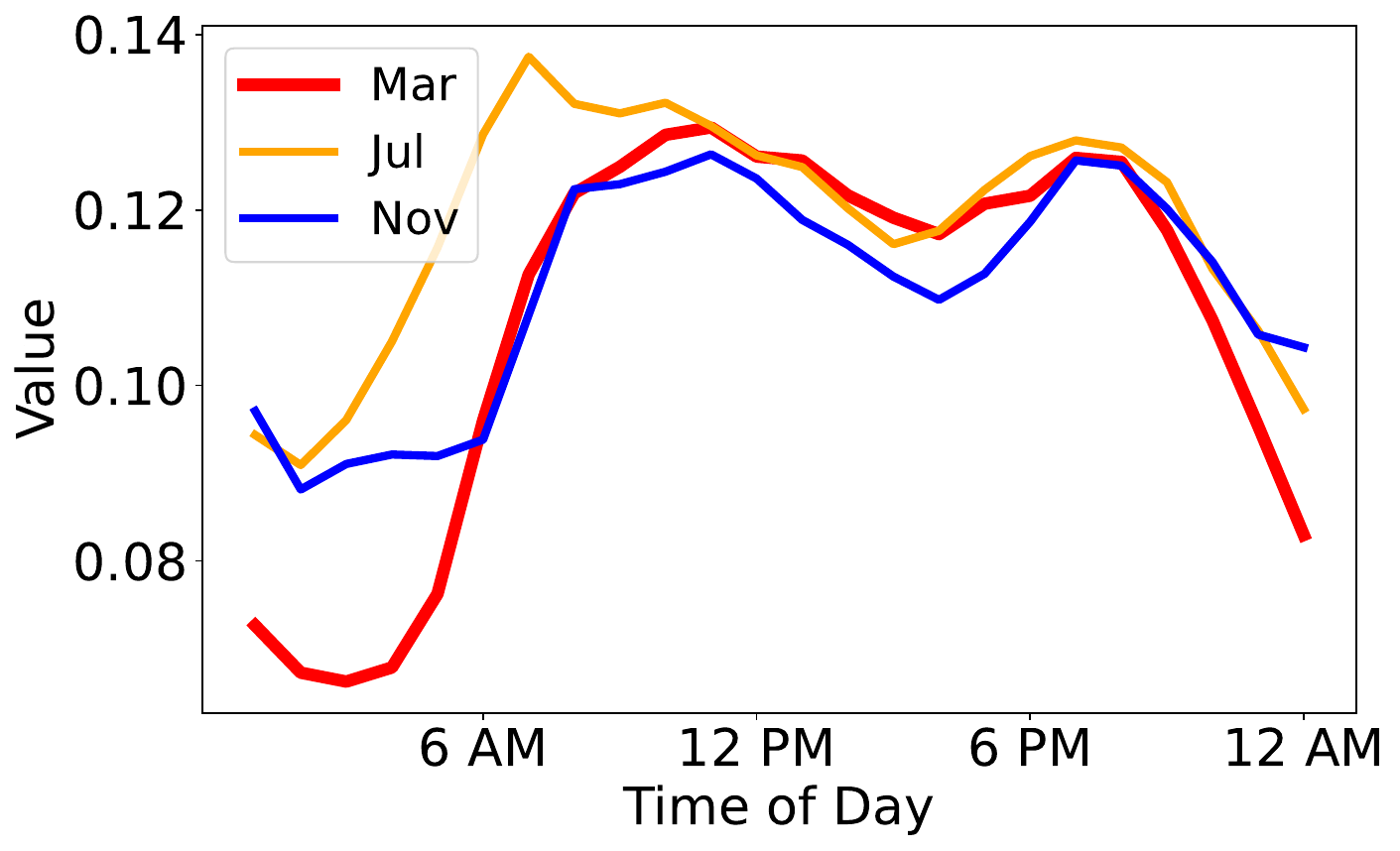}
    \caption{Water}
    \label{fig:sub3}
\end{subfigure}
\caption{Aggregated statistics of each whole utility dataset on March, July, and November. The solid line represents the global mean of each data point in the time series, while the shaded area indicates the range between the 25th and 75th percentiles for the electricity time series. Due to the sparsity of data points in the gas and water datasets, we omit the quantile plots for these datasets. }
\label{fig:agg_dataset}
\end{figure*}   
\begin{figure*}[]
\centering
\begin{subfigure}[b]{0.33\linewidth}
    \includegraphics[width=\linewidth]{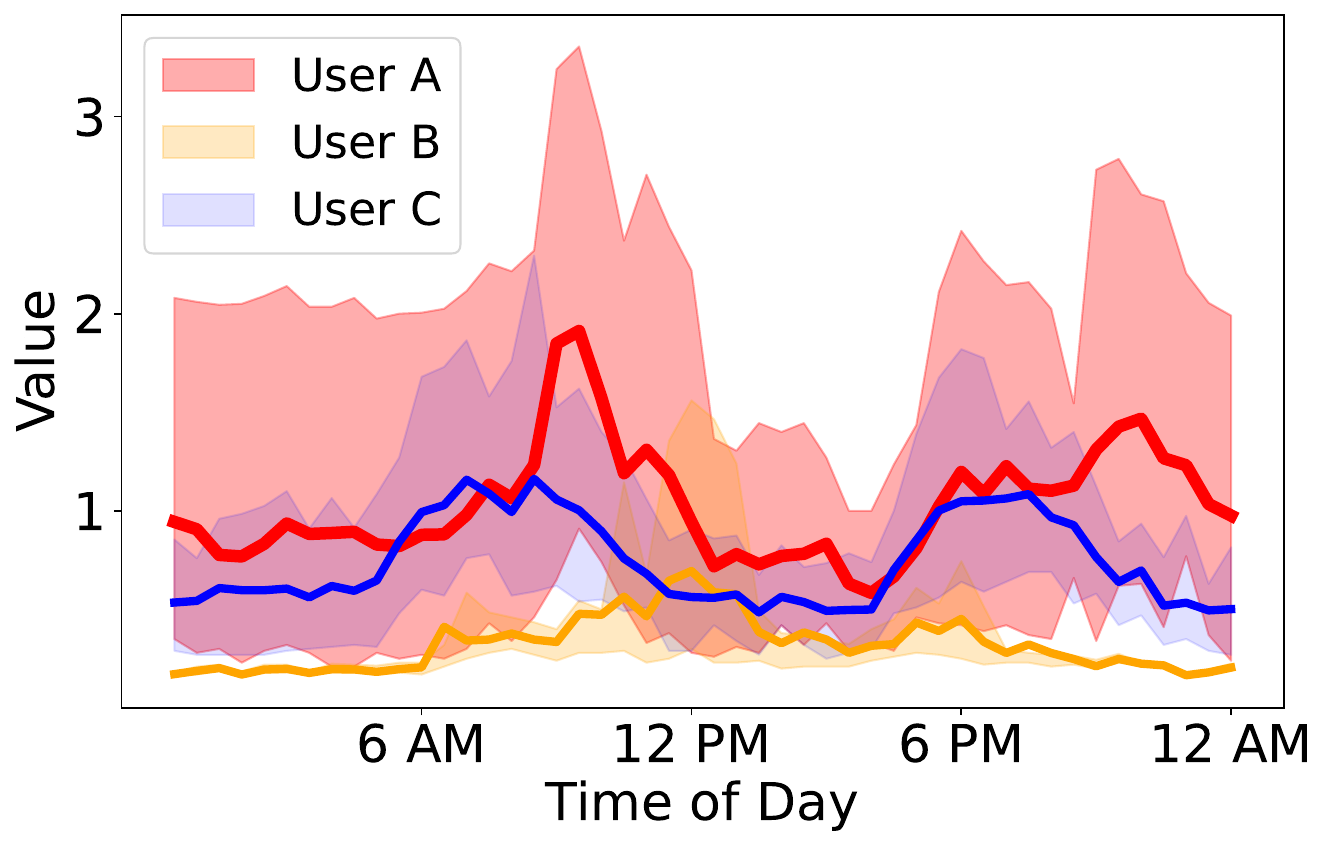}
    \caption{2018/03}
    \label{fig:sub1}
\end{subfigure}
\hfill
\begin{subfigure}[b]{0.33\linewidth}
    \includegraphics[width=\linewidth]{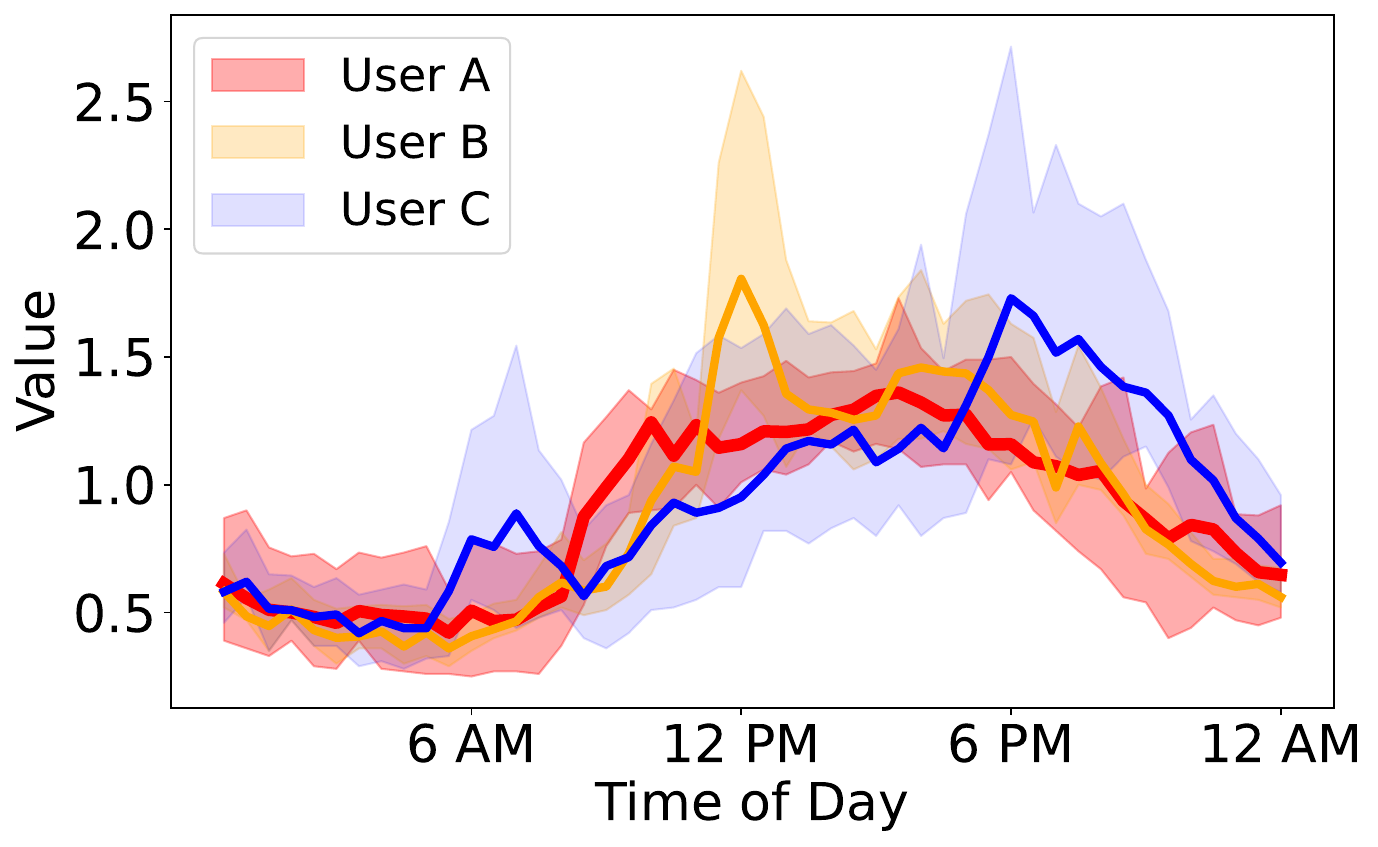}
    \caption{2018/07}
    \label{fig:sub2}
\end{subfigure}
\hfill
\begin{subfigure}[b]{0.33\linewidth}
    \includegraphics[width=\linewidth]{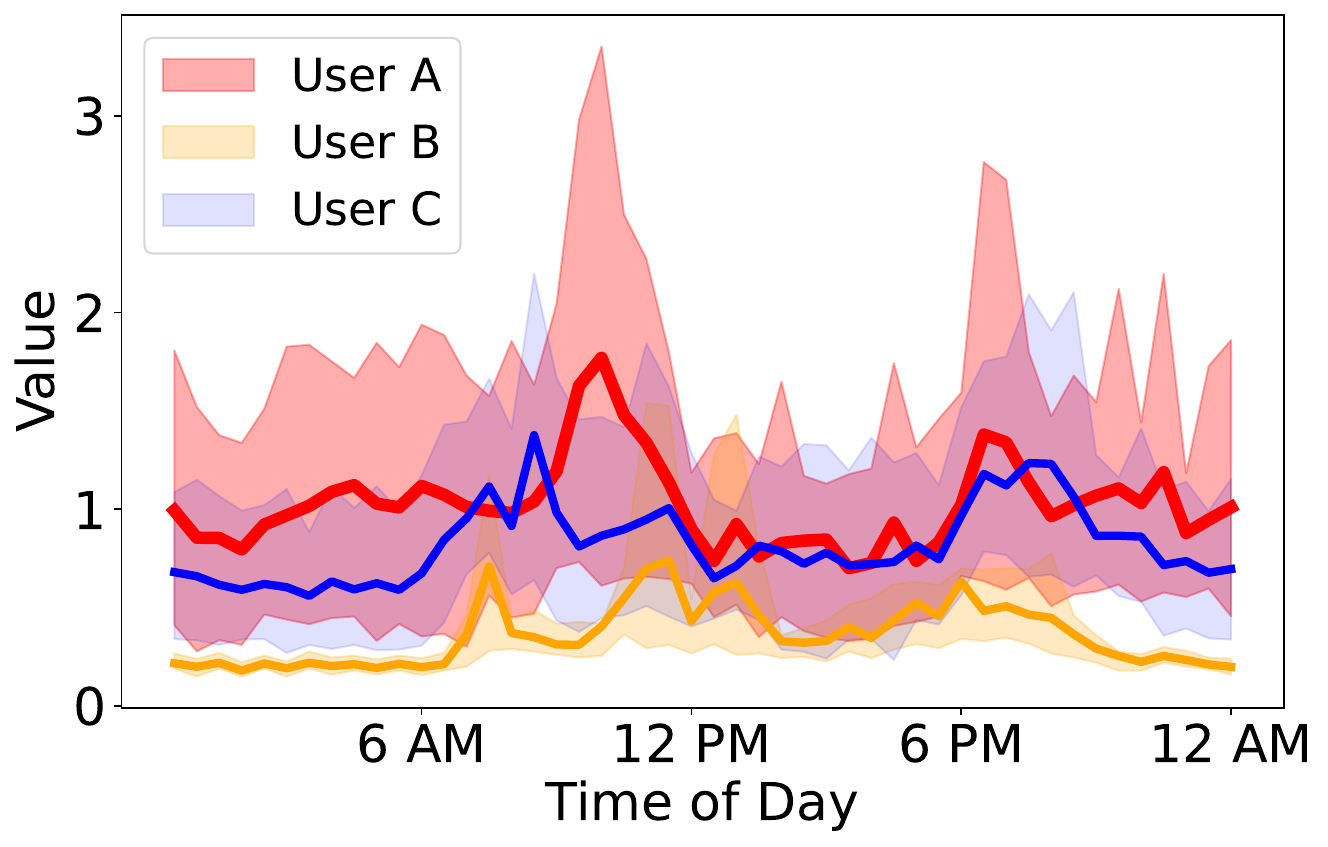}
    \caption{2018/11}
    \label{fig:sub3}
\end{subfigure}
\caption{Aggregated statistics of each user in different months on electricity dataset.}
\label{fig:user_dataset5}
\end{figure*}
\begin{figure*}[]
\centering
\begin{subfigure}[b]{0.33\linewidth}
    \includegraphics[width=\linewidth]{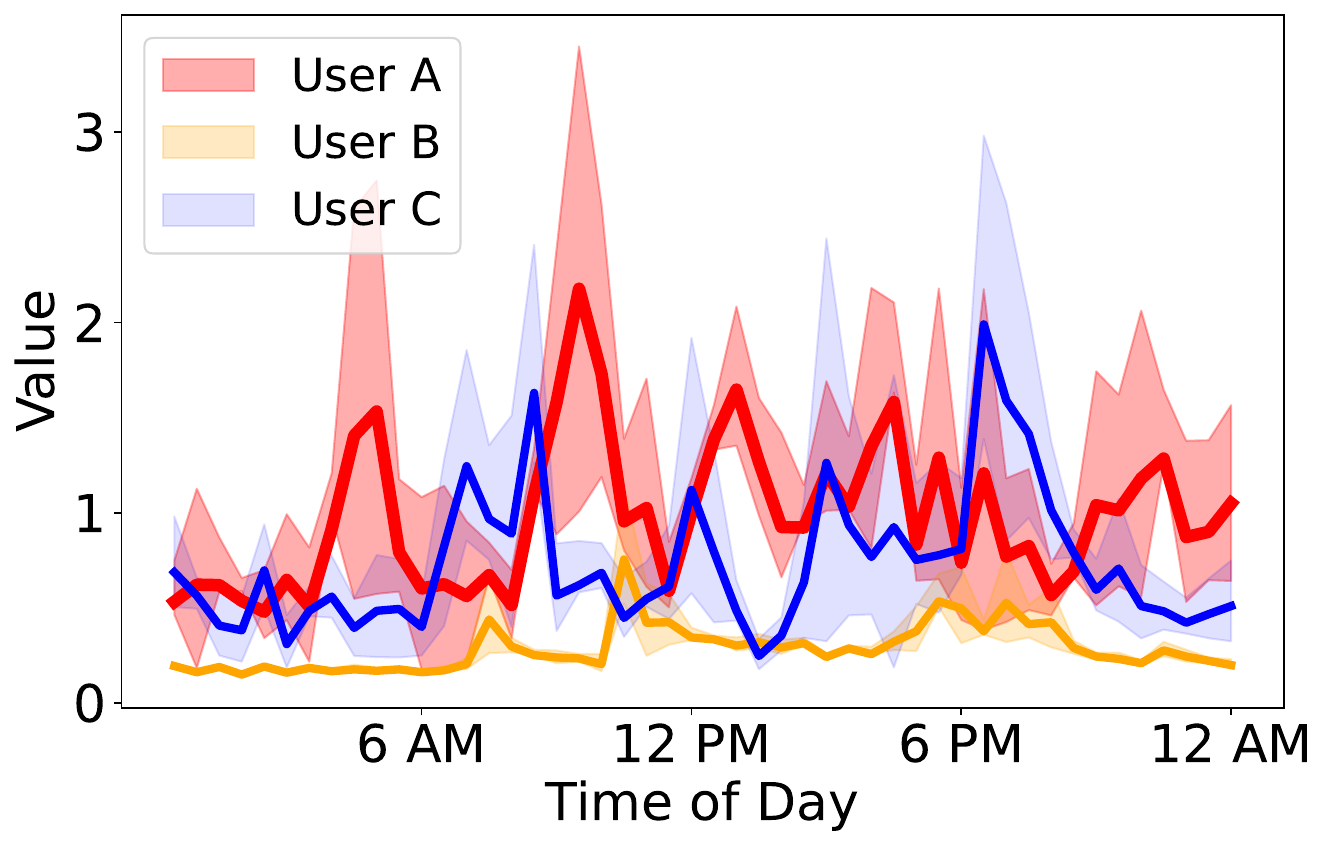}
    \caption{Monday}
    \label{fig:sub1}
\end{subfigure}
\hfill
\begin{subfigure}[b]{0.33\linewidth}
    \includegraphics[width=\linewidth]{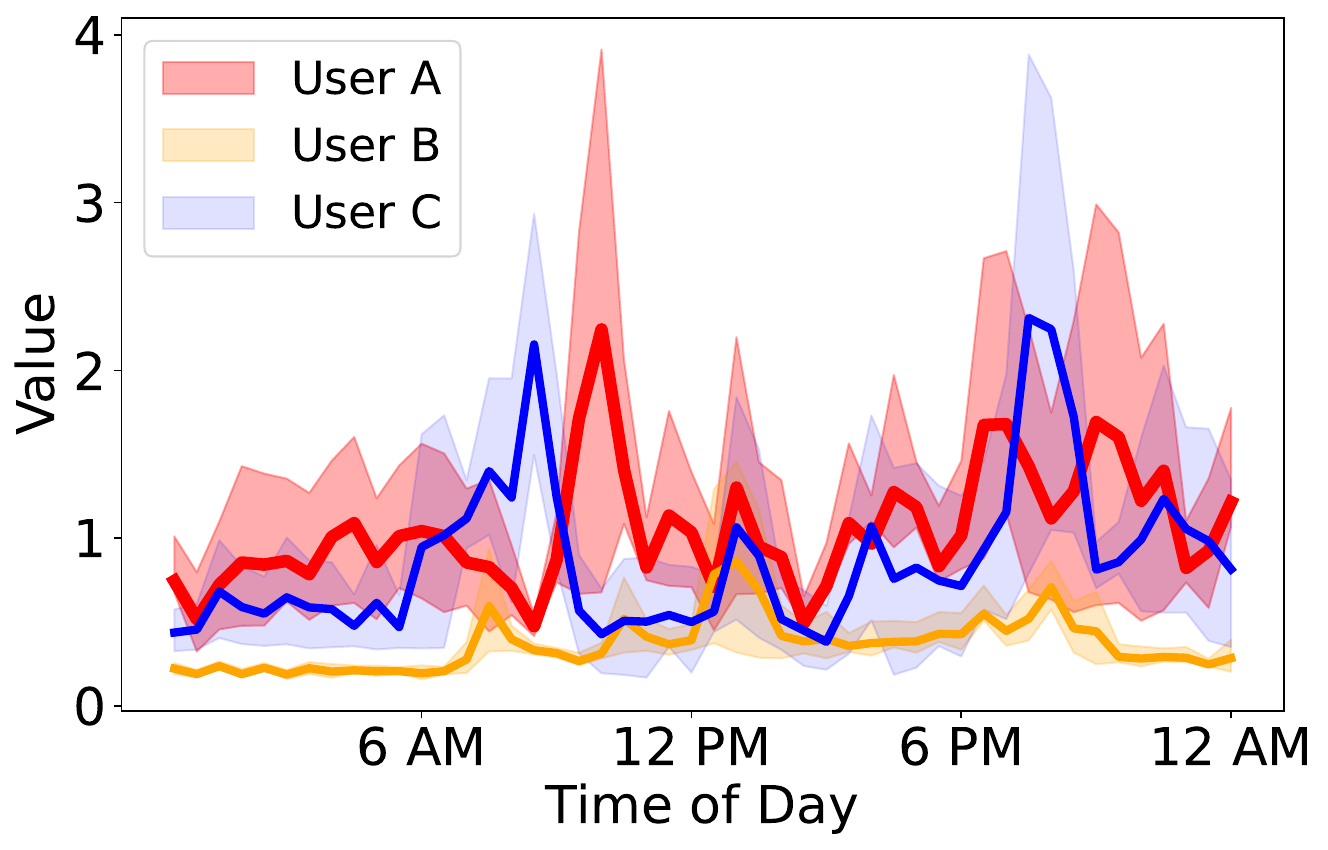}
    \caption{Tuesday}
    \label{fig:sub2}
\end{subfigure}
\hfill
\begin{subfigure}[b]{0.33\linewidth}
    \includegraphics[width=\linewidth]{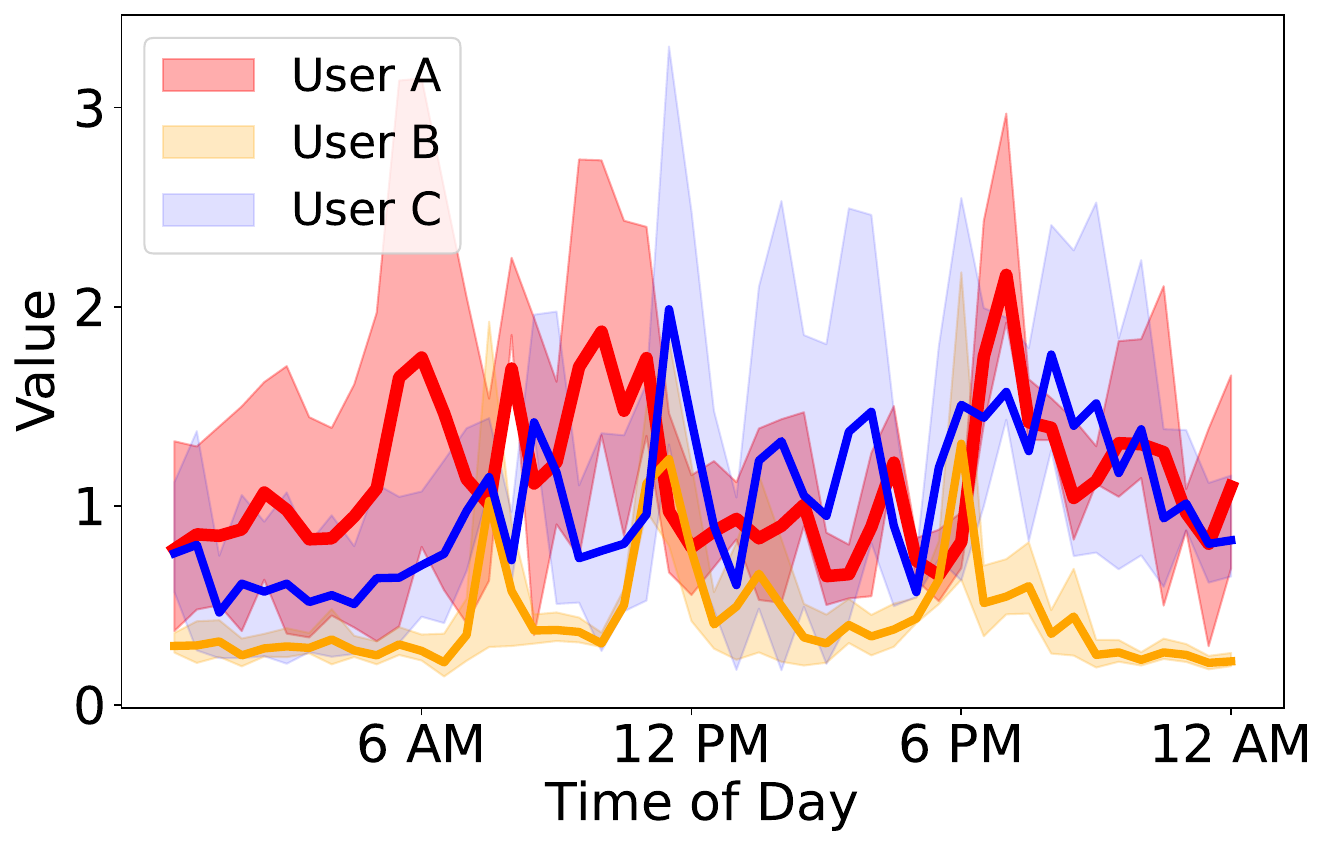}
    \caption{Wednesday}
    \label{fig:sub3}
\end{subfigure}
\\
\begin{subfigure}[b]{0.33\linewidth}
    \includegraphics[width=\linewidth]{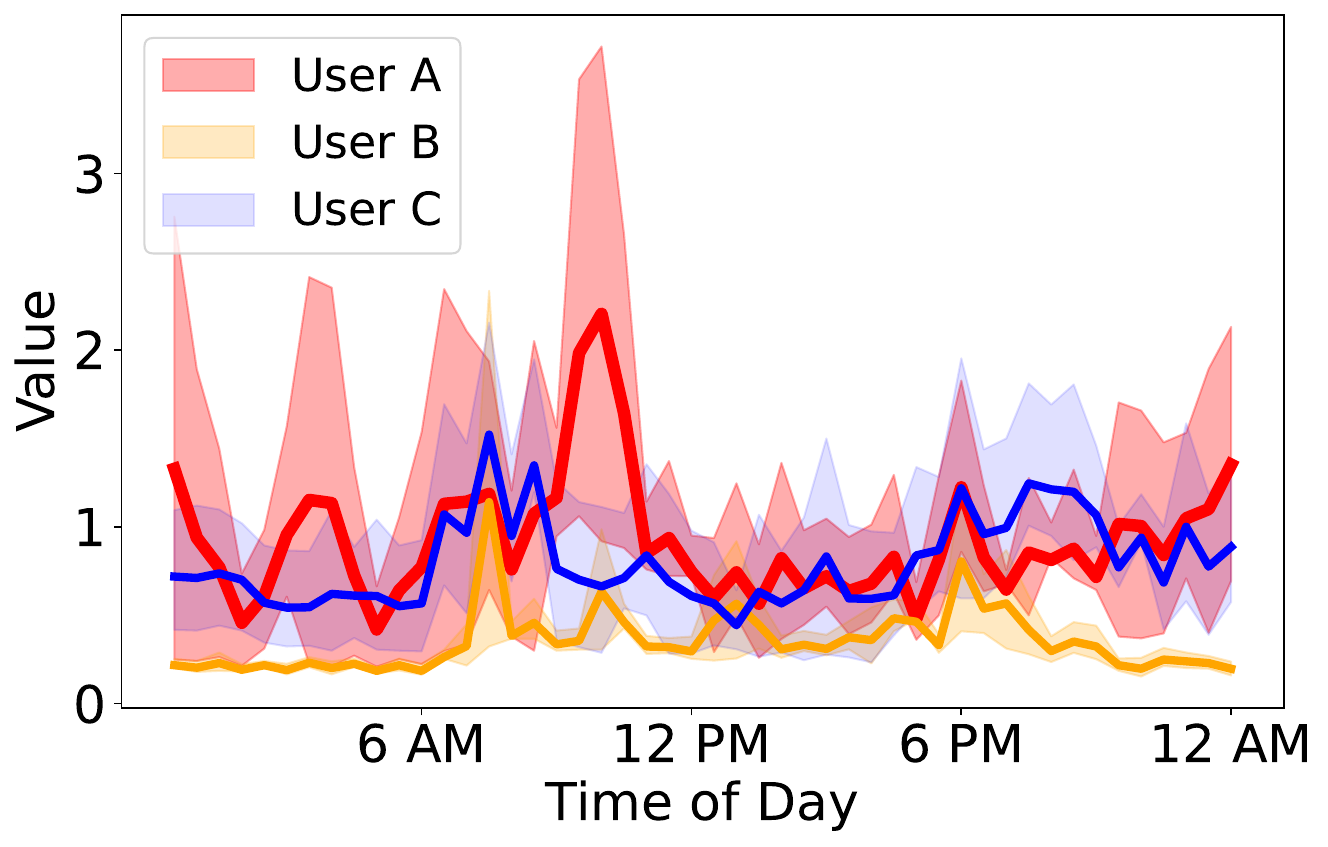}
    \caption{Thursday}
    \label{fig:sub4}
\end{subfigure}
\hfill
\begin{subfigure}[b]{0.33\linewidth}
    \includegraphics[width=\linewidth]{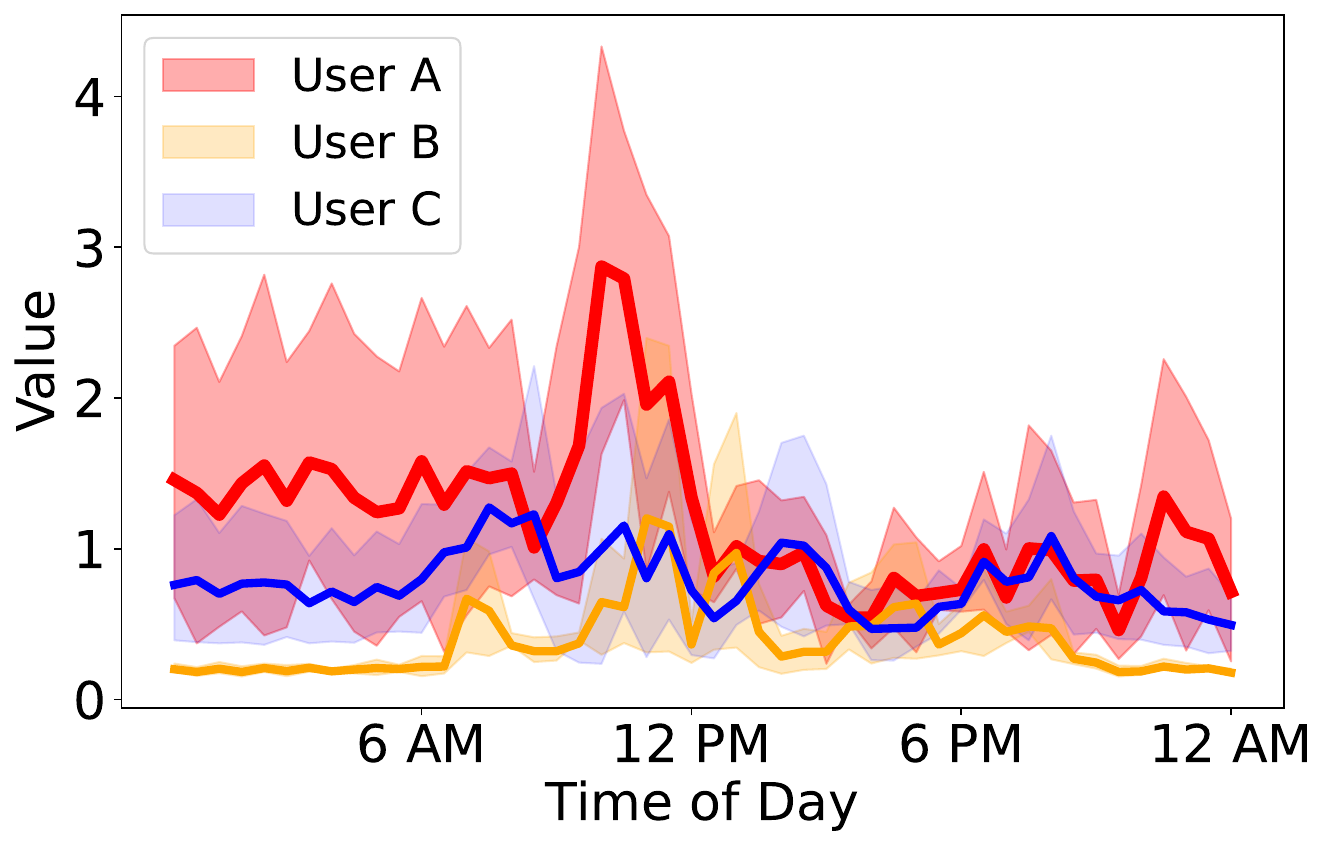}
    \caption{Friday}
    \label{fig:sub5}
\end{subfigure}
\hfill
\begin{subfigure}[b]{0.33\linewidth}
    \includegraphics[width=\linewidth]{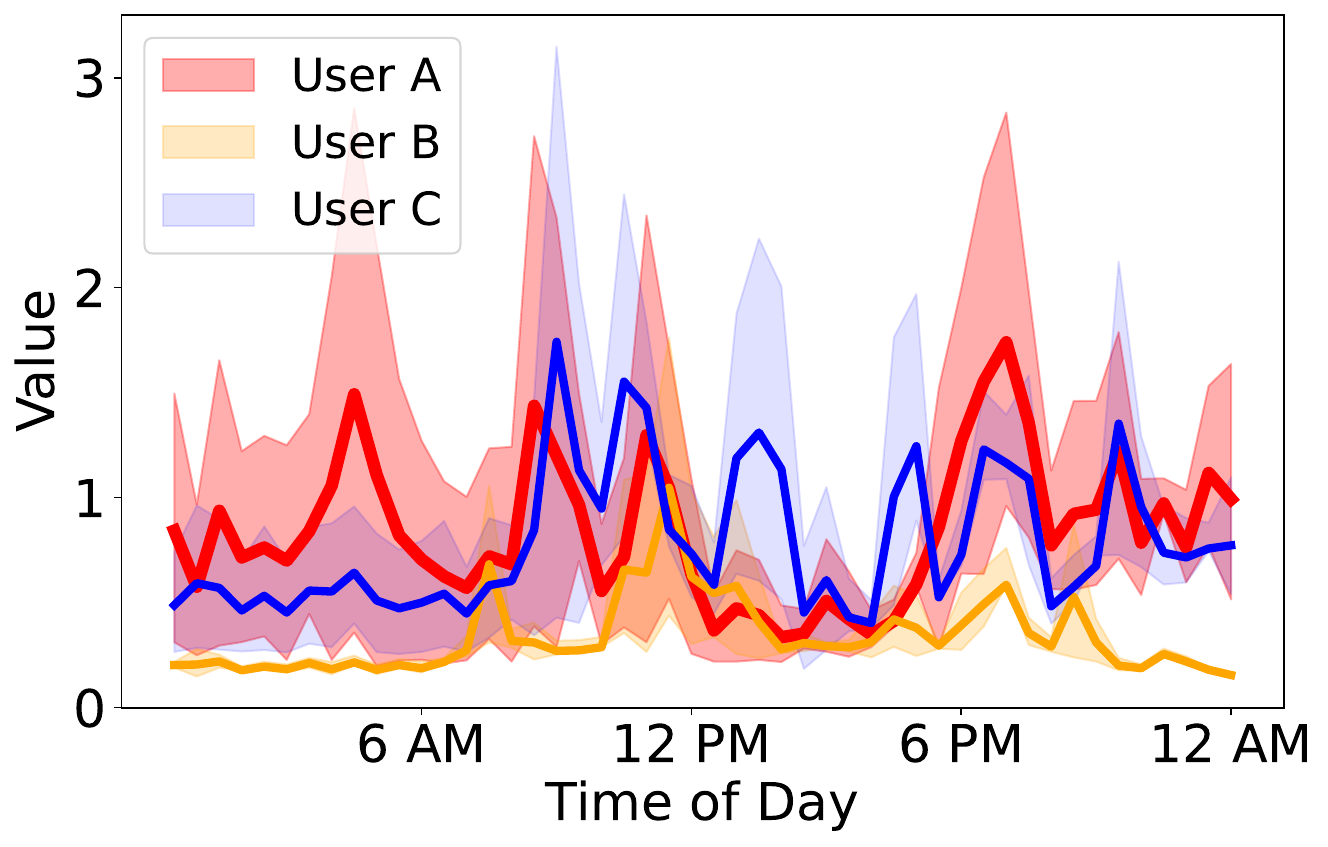}
    \caption{Saturday}
    \label{fig:sub6}
\end{subfigure}
\caption{Aggregated statistics of each user on different days in a week on the electricity dataset.}
\label{fig:user_dataset}
\end{figure*}

We present some analysis and visualization of the datasets in this part.
As shown in Fig. \ref{fig:agg_dataset}, 
daily usage patterns of different utilities are distinct and each utility usage also varies significantly across different months. For electricity consumption, March and November show almost the same pattern, but July shows notably higher consumption, especially between 10 AM and 10 PM, which is likely due to increasing air conditioner usage in the hot summer in Florida. The usage patterns of gas and electricity consumption are opposite. Gas consumption in July is much less than that in other months.
Water consumption shows consistent patterns across all three months and there is always a morning peak and an evening peak. This regular pattern stays stable from March to November, though July has a slightly higher daytime usage.

Fig.~\ref{fig:user_dataset5} illustrates individual-level electricity consumption patterns for three randomly chosen users in March, July, and November. It indicates that different users exhibit distinct energy consumption behaviors. During these three months, user C consistently shows higher electricity usage compared to user A and user B, with broader fluctuations throughout the day, while user A and user B show more steady consumption patterns. Even for the same user, the electricity consumption varies across different months. User A maintains relatively low electricity usage throughout these three months with minimum fluctuations throughout the day. User B consumes similar electricity to user A in March and July, but the consumption is much higher in November compared to user A. Overall, this analysis demonstrates the impact of both user-specific behaviors and ambient factors like seasonal changes on electricity consumption patterns.

Fig. \ref{fig:user_dataset} shows the electricity consumption pattern of three randomly selected users across different days of the week, revealing significant variations in daily usage behaviors. During weekdays, user A consistently consumes the most electricity compared to user B and user C, with several peaks during the day, especially around 8 AM to 10 AM and 6 PM to 9 PM, which aligns with our previous monthly analysis. Unlike user A, user B keeps their usage low and steady, and user C shows moderate usage with some ups and downs. There are the highest peaks on Monday, especially for user A, while there are longer periods of high electricity consumption on Thursday and Friday. On Saturday, all users show similar electricity usage patterns with two main peaks: one around 1 PM, which is likely during lunchtime, and another one around 8 PM, which is likely for evening activities. This similar pattern indicates that people's activity is more similar on weekends, unlike weekdays, where individual schedules create more distinct usage patterns.

\clearpage
\subsection{Extended Sensitivity Analysis}
\label{app:sensitivity}
We report the complete sensitivity analysis results of all methods on the electricity dataset under point and block missingness at 20\%, 35\%, and 50\% missing rates. Table~\ref{tab:s_crps} and Table~\ref{tab:s_mae&mse} summarize the CRPS and the MAE/MSE results, respectively.
\begin{table}[H]
\small
\caption{Sensitivity Analysis: The results of CRPS for electricity data imputation under different missing ratios.}\vspace{-5pt}
\label{tab:s_crps}
\setlength{\tabcolsep}{2.7pt}
\begin{tabular}{lcccccc}
\toprule
\textbf{Method} & \multicolumn{2}{c}{\textbf{20\%}} & \multicolumn{2}{c}{\textbf{35\%}} & \multicolumn{2}{c}{\textbf{50\%}}\\
\cmidrule(lr){2-3} \cmidrule(lr){4-5} \cmidrule(lr){6-7}
\textbf{Missing} & Point & Block  & Point  & Block  & Point  & Block  \\ 
\midrule
GP-VAE      &0.0292 &\textbf{0.0300 }&0.0420 &0.0363 &0.0415 &0.0611   \\
CSDI        &0.0233 &0.0377 &0.0411 &\underline{0.0394} &0.0419 &0.0528   \\
PriSTI      &0.0225 &0.0389 &\underline{0.0387} &0.0399 &\textbf{0.0396} &\underline{0.0519}   \\ 
ImputeFormer   &0.1347 &0.3328 &0.1480 &0.3595 &0.1567 &0.3851   \\
TimeMixer++    &\textbf{0.0199} &0.1561 &0.0983 &0.1521 &0.1804 &0.1924 \\
\midrule
\modelAbvName &\underline{0.0210} &\underline{0.0359} &\textbf{0.0374} &\textbf{0.0386} &\underline{0.0403} &\textbf{0.0479}   \\
\bottomrule
\end{tabular}
\end{table}

\begin{table*}[b]
\centering
\small
\caption{Sensitivity Analysis: The results of MAE and MSE for electricity data imputation under different missing ratios.}
\label{tab:s_mae&mse}\vspace{-5pt}
\begin{tabular}{lcccccccccccc}
\toprule
& \multicolumn{4}{c}{\textbf{20\%}} & \multicolumn{4}{c}{\textbf{35\%}} & \multicolumn{4}{c}{\textbf{50\%}} \\
\cmidrule(lr){2-5} \cmidrule(lr){6-9} \cmidrule(lr){10-13}
\textbf{Method} & \multicolumn{2}{c}{Point Missing} & \multicolumn{2}{c}{Block Missing} & \multicolumn{2}{c}{Point Missing} & \multicolumn{2}{c}{Block Missing} & \multicolumn{2}{c}{Point Missing} & \multicolumn{2}{c}{Block Missing} \\
\cmidrule(lr){2-3} \cmidrule(lr){4-5} \cmidrule(lr){6-7} \cmidrule(lr){8-9} \cmidrule(lr){10-11} \cmidrule(lr){12-13}
& MAE & MSE & MAE & MSE & MAE & MSE & MAE & MSE & MAE & MSE & MAE & MSE \\ 
\midrule
G-Mean      &0.5021 &0.4530 &0.5062 &0.4781 &0.5069 &0.4695 &0.5126 &0.4744 &0.5064 &0.4708 &0.4991 &0.4584 \\
U-Mean      &0.3373 &0.3045 &0.3813 &0.3176 &0.3750 &0.3050 &0.3822 &0.3140 &0.3788 &0.3044 &0.3732 &0.3021 \\
LINEAR      &0.2130 &0.1255 &0.2706 &0.2115 &0.2659 &0.2008 &0.3290 &0.2950 &0.3639 &0.3568 &0.3841 &0.4281 \\
KF          &0.3286 &0.5099 &0.5020 &0.7879 &0.5012 &0.8765 &0.8000 &2.5386 &0.8968 &3.2182 &1.4544 &9.7296 \\
SAITS       &0.2170 &0.0141 &0.2019 &0.0438 &0.1587 &0.0598 &0.2946 &0.0917 &0.1686 &0.0368 &0.1532 &0.0367 \\
GP-VAE      &0.0382 &0.0031 &0.0383 &0.0034 &0.0523 &0.0049 &0.0459 &0.0052 &0.0522 &0.0068 &0.0818 &\underline{0.0141} \\
CSDI        &0.0293 &\underline{0.0025} &0.0301 &\underline{0.0031} &0.0514 &0.0055 &0.0423 &\textbf{0.0038} &0.0519 &0.0059 &0.0659 &0.0153 \\
PriSTI      &0.0288 &0.0031 &0.0326 &0.0034 &\underline{0.0402} &\textbf{0.0043} &\underline{0.0388} &0.0039 &\underline{0.0472} &\underline{0.0052} &\textbf{0.0514} &\textbf{0.0129} \\
TimeMixer++ &\textbf{0.0199} &0.0031 &0.1561 &0.1368 &0.0983 &0.0565 &0.1521 &0.1427 &0.1804 &0.1583 &0.1924 &0.1621 \\
ImputeFormer &0.1347 &0.0952 &0.3328 &0.1252 &0.1480 &0.1025 &0.3595 &0.2235 &0.1567 &0.1236 &0.3851 &0.2568 \\
\midrule
\modelAbvName &\underline{0.0276} &\textbf{0.0024} &\underline{0.0293} &\textbf{0.0028} &\textbf{0.0387} &\underline{0.0049} &\textbf{0.0346} &\underline{0.0039} &\textbf{0.0396} &\textbf{0.0050} &\underline{0.0539} &0.0149 \\
\bottomrule
\end{tabular}
\end{table*}

\end{document}